%% file: main.tex
\documentclass[12pt]{article}  
\usepackage[a4paper, margin=1.2in]{geometry}
\usepackage{graphicx}  
\usepackage{amsmath}   
\usepackage{amssymb}   
\usepackage{cite}      
\usepackage{tikz}
\usepackage{float}
\usepackage{caption}
\usepackage{subcaption}
\usepackage{mathtools}
\usepackage{bm}
\usepackage{tablefootnote}
\usetikzlibrary{decorations.pathreplacing, 
decorations.pathmorphing, 
positioning, 
arrows, 
fit, 
backgrounds}
\usepackage{placeins}

\usepackage[hidelinks]{hyperref} 
\numberwithin{equation}{section} 

\newcommand{\id}{\mathbb{I}}
\title{Spectral Architecture Search for Neural Network Models}
\author{\normalsize Gianluca Peri\thanks{Department of Physics and Astronomy, University of Florence, Sesto Fiorentino, Italy} \and \normalsize Lorenzo Chicchi\footnotemark[1] \and \normalsize Duccio Fanelli\footnotemark[1] \and \normalsize Lorenzo Giambagli\thanks{Department of Physics, Freie Universit\"at Berlin, Arnimallee 12, 14195, Berlin, Germany}}

\date{}  
\begin{document}
	
	\maketitle
	
\begin{abstract}
Architecture design and optimization are challenging problems in the field of artificial neural networks. Working in this context, we here present SPARCS (SPectral ARchiteCture Search), a novel architecture search protocol which exploits the spectral attributes of the inter-layer transfer matrices. SPARCS allows one to explore the space of possible architectures by spanning continuous and differentiable manifolds, thus enabling for gradient-based optimization algorithms to be eventually employed. With reference to simple benchmark models, we show that the newly proposed method yields a self-emerging architecture with a minimal degree of expressivity to handle the task under investigation and with a reduced parameter count as compared to other viable alternatives.
\end{abstract}
	

\section{Introduction}

Neural networks are very effective machine learning tools that prove extremely valuable in unwinding the best representation of the data at hand. To improve the ability of neural networks to automatically perform the assigned tasks, innovative architectures have been proposed and thoroughly tested. Employed architectures have been customarily developed by human experts, with manual, time-consuming, and error-prone processes. To go beyond manual design, novel algorithmic strategies for automated discovery of optimal neural architectures have been developed. Consequently, architecture engineering has become a relevant field of active research \cite{JMLR:v20:18-598,
10.1117/12.2613628}. 

Neural Architecture Search (NAS), the process that seeks to optimize
network architecture, has been successfully applied on tasks as image classification \cite{8579005, Real_Aggarwal_Huang_Le_2019}, object detection \cite{8579005}, or semantic segmentation \cite{NEURIPS2018_c90070e1}, yielding remarkable performance, as compared to manually designed benchmarks. According to \cite{JMLR:v20:18-598}, NAS is a subfield of Automated Machine Learning (AutoML) \cite{hutter2019automated}, the process that aims at automating the steps  propaedeutic to applying machine learning to real-world problems. It also shows a notable overlap with hyperparameter optimization (a critical process in machine learning that involves selecting the optimal set of hyperparameters for a learning algorithm \cite{feurer_hyperparameter_2019}) and meta-learning (a subfield of machine learning that focuses on training models to understand and adapt to new tasks autonomously \cite{vanschoren_meta_2019}). 

 Starting from these premises, we here propose SPARCS (SPectral ARchiteCture Search), a training algorithm for deep neural networks 
 which builds on the spectral parametrization of the inter-layer transfer 
 matrices to automatically identify the  best possible architecture given the specific problem under scrutiny. To this end, the multi-layered network is initially set to operate as a perceptron: only direct connections from the input nodes towards the destination layer are hence active. Upon training, and depending on the inherent complexity of the examined problem, hidden layers get recruited, via feedforward or skip (long ranged) connections, and thus contribute to the performed computation. As we shall prove, this yields a rather compact network, in terms of links activated {\it ex post}, as compared to viable approaches rooted on a conventional (non spectral) formulation of the training problem. Importantly, SPARCS explores the space of possible architectures by scanning a continuous and differentiable manifold, with gradient-based optimization algorithms. In this work, we set up the theory of SPARCS for a neural network of arbitrary dimensions and validate its adequacy against simple benchmark applications. The first benchmark aims at testing the efficacy of our model in grasping the linear to non-linear transition of a dataset, by adequately adapting the architecture. The second deals with a regression problem of a randomly generated neural network, with weight distributions which can be approximated with just one hidden layer. This  solution is indeed correctly spotted by our method. Additional tests performed by operating SPARCS in conjunction with Convolutional Neural Networks for image classification tasks (CIFAR-10 and CIFAR-100) are reported in Appendix \ref{AppCIFAR} and commented upon in the main body of the paper. The hyperparameters employed in carrying out the above tests are summarized for convenience in a set of Tables reported in Appendix \ref{hyperparameters}.
 
Historically, NAS methods leveraged evolutionary algorithms to explore the architecture's space. One famous example is NEAT (\textit{Neuro-Evolution of Augmenting Topologies}) \cite{stanley:ec02}. This method encodes the topology of the network in a set of genes, then a population of networks is created and put under evolutionary pressure. The fitness of a network is given by the displayed performance on the examined task (classification, regression, etc.). This can be seen as a sort of hyperparameter optimization process, where the search space of parameters is explored through evolutionary means. Another family of methods reformulate NAS  as a veritable hyperparameter search problem. These are the so called RL-NAS (\textit{Reinforcement Learning based Neural Architecture Search}). These latter methods use a neural network to generate a neural network architecture, then train and test an instance of such architecture on the task at hand, and finally use the recorded performance as a reward signal to feed the generating network scheme\cite{zoph2017neuralarchitecturesearchreinforcement, zoph2018learningtransferablearchitecturesscalable}. These two families of methods are still used and researched today, however, with time, performance considerations pushed the community towards adopting a \textit{differentiable architecture search} approach:  during training, the optimization of both the architecture and the model's weights are simultaneously carried out. One famous example is DARTS (\textit{Differentiable ARchiTectures Search}). The original algorithm resulted in popular variants such as PC-DARTS, GDAS, FairDARTS \cite{liu2019dartsdifferentiablearchitecturesearch, xu2020pcdartspartialchannelconnections, dong2019searchingrobustneuralarchitecture, chu2020fairdartseliminatingunfair}. The main idea behind all of these methods is to associate continuous parameters to the topological features of the network, and then  optimize them, along with the network's weights, through iterative procedures, like \textit{gradient descent}. In this context SPARCS places itself as a new differentiable NAS method, but with three notable addition that we anticipate hereafter:

\begin{itemize}
\item  SPARCS adjustable parameters are the eigenvalues and the eigenvectors of the network's adjacency matrix, which naturally reflect on both the weights and the architecture of the ensuing network.
\item under SPARCS, the network can be initialized by imposing a minimal architectural backbone, supplemented with a form of regularization on the architecture extension: as we shall see, the proposed algorithm can self-consistently discover small architectures to handle the supplied tasks.
\item The spectral parametrization of SPARCS naturally leads to a form of parameter sharing between different structures of the network: this allows to perform architecture search with a small parameter count, as compared to non-spectral alternatives.
\end{itemize}

As an additional remark, we stress that the above referenced NAS methods prove useful when the training for the specific problem under scrutiny requires a limited amount of time,  in such a way that multiple training attempts can be tolerated at the price of a reasonable computation cost. Unfortunately that is not always the case. 
If the training takes several weeks on powerful computing resources, the usage of methods like NEAT and DARTS, which come along multiple training sessions, appear prohibitive, at least on standard hardware. SPARCS can, at least in principle, fill this gap as (i) the architecture search is performed during the main (and sole) training session, with no need for multiple runs; (ii) the number of trainable parameters is set by the largest architecture in the explored example. More into details, SPARCS exploits a non trivial generalization of the so called {\it spectral parametrization} \cite{spectral_learning,
teacher_student_giambagli, SpectralPrune, preChicchi}, a recently introduced scheme for handling the optimization of feedforward networks. In a nutshell, feedforward neural networks can be pictured as an ordered sequence of linear transformations (interspersed by non linear filters), each encoded in a square adjacency matrices for the ensuing bipartite directed graph. The spectral paradigm reformulates the optimization process in dual space, the eigenvalues and the eigenvectors of the 
inter-layer adjacency matrix, supplemented with the inclusion of self-loops, acting as target free parameters. Notice that the spectral approach to training requires no costly spectral decomposition step, as the network is formulated from the outset in terms of eigenvectors and eigenvalues. To be able to operate Neural Architecture Search in the spectral domain, one needs first to extend the realm of validity of the spectral parametrization beyond the simple entry setting, where the information transfer between two consecutive layers is solely considered. Working along these lines, we will here set  the mathematical foundation for a spectral characterization for the multi-layered interactions between an arbitrary collection of inter-tangled stacks, of different sizes and subject to local and long-ranged couplings. This generalized framework prompts to a wider usage of the  spectral technology, beyond what so far explored in the literature and aside the specific application here discussed. 

Recapitulating one more time, we focus first on extending the validity of the spectral parametrization beyond the simple case of two consecutive layers, so as to account for the general setting where an arbitrary number of layers is accounted for. As an immediate application of the proposed framework,  we present and thoroughly test an innovative scheme for automated neural network architecture search. The algorithm initializes the networks as simple perceptrons. Then, it adds additional complexity via hidden layers based on task demands, with a parsimonious approach which tends to favor the simplest solution possible, given the problems constraints. This results in efficient computational models with fewer parameters, as compared to other viable alternatives. Our conclusions are supported by a campaign of tests that we have performed on benchmark models. To contribute with a novel formulation of the optimization problem (which generalizes the existing spectral perspective) and to provide a novel scheme for architecture search (which exploits the proposed mathematical framework) define the main motivations of our study.

The paper is organized as follows: in the next Section we define the reference background and set the mathematical framework. Then, we move forward to discussing the generalized spectral theory for a simple network with just one hidden layer. In the successive Section we focus instead on deep neural networks of arbitrary size, before turning to discuss SPARCS and its application to our tests frameworks. Finally, we sum up and conclude. Relevant calculations and additional numerical tests are reported in the annexed Appendices.
 
 \section{Mathematical background: the spectral theory for a feed-forward network with local connections}   

Consider a deep feedforward network made of $\ell$ distinct layers and label each layer with the index $i$ $(=1,...,\ell)$. Denote by  $N_i$ the number of neurons which belong to the $i-$th layer. The total number of parameters that one seeks to optimise when dealing with a typical fully connected neural network (all neurons of any given layer with $i < \ell-1$ are linked to every neurons of the adjacent layer) equals $\sum_{i=1}^{\ell-1} N_i N_{i+1}$, when omitting additional bias. At this level, no skip connections, namely long-ranged links between distant (i.e. non adjacent) layers are considered. Working within this setting, we will review in the following the core idea of the spectral parametrization as introduced in 
\cite{spectral_learning}. This prelude is meant to provide the necessary background for the non trivial extension hereafter addressed.

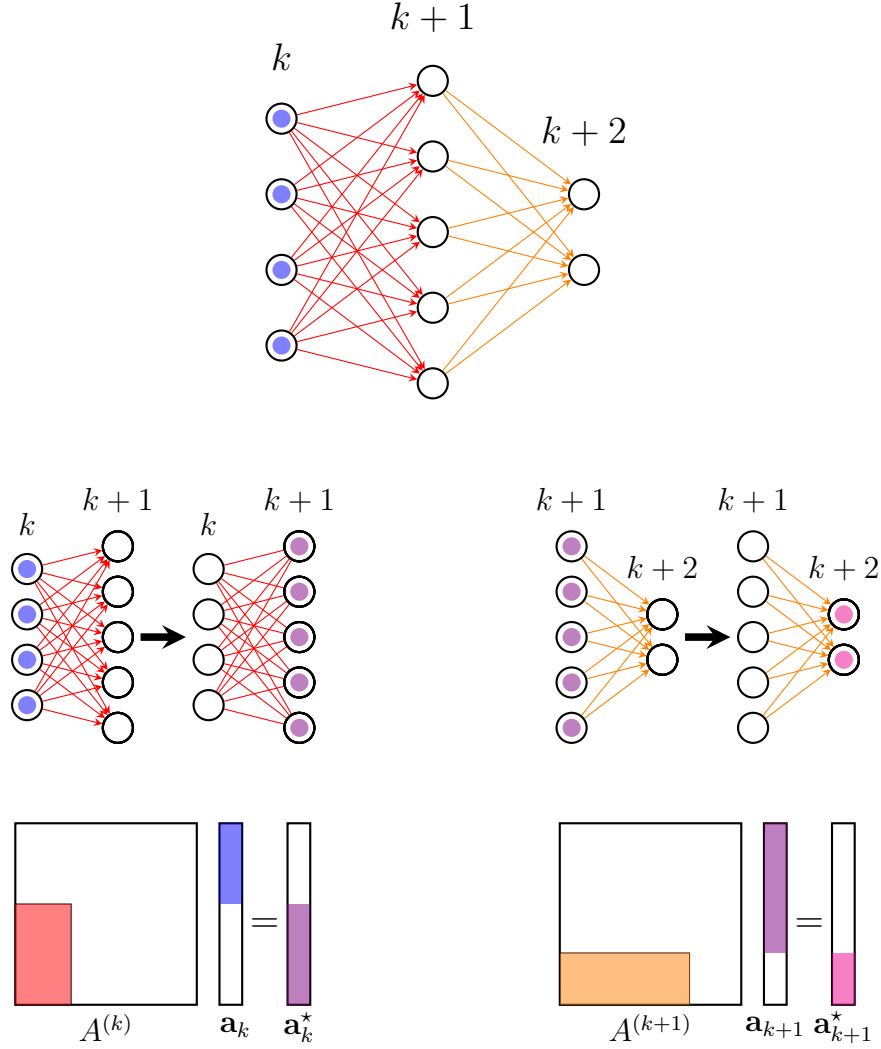
\begin{figure}[ht]
    \centering
    
    \begin{tikzpicture}[>=stealth]
        \foreach \i in {1,...,4} {
            \node[circle,draw,inner sep=4pt, thick] (l0\i) at (1,-\i) {};
            \fill[blue, opacity=0.5] (l0\i) circle (0.12);
        }

        \foreach \i in {1,...,5} {
            \node[circle,draw,inner sep=4pt, thick] (l1\i) at (3,-\i+0.5) {};
        }

        \foreach \i in {1,...,2} {
            \node[circle,draw,inner sep=4pt, thick] (l2\i) at (5,-\i-1) {};
        }

        \foreach \i in {1,...,4} {
            \foreach \j in {1,...,5} { 
                \draw[red, ->] (l0\i) -- (l1\j);
            }
        }

        \foreach \i in {1,...,5} {
            \foreach \j in {1,...,2} {
                \draw[orange, ->] (l1\i) -- (l2\j);
            }
        }

        \node[above=0.5cm] at (l01) {\large$k$};
        \node[above=0.5cm] at (l11) {\large$k+1$};
        \node[above=0.5cm] at (l21) {\large$k+2$};
    \end{tikzpicture}
    
    \vspace{1cm}

    \begin{tikzpicture}[>=stealth, scale=0.6]
        \foreach \i in {1,...,4} {
            \node[circle,draw,inner sep=4pt, thick] (h\i) at (-9,-\i-0.5) {};
            \fill[blue, opacity=0.5] (h\i) circle (0.2);
        }
        
        \foreach \i in {1,...,4} {
            \foreach \j in {1,...,5} {
                \node[circle,draw,inner sep=4pt, thick] (out\j) at (-7,-\j) {};
                \draw[red, ->] (h\i) -- (out\j);
            }
        }

        \draw[->, line width=1mm] (-6.5,-3) -- (-5.5,-3);

        \foreach \i in {1,...,4} {
            \node[circle,draw,inner sep=4pt, thick] (h2\i) at (-5,-\i-0.5) {};
        }
        
        \foreach \i in {1,...,4} {
            \foreach \j in {1,...,5} {
                \node[circle,draw,inner sep=4pt, thick] (out2\j) at (-3,-\j) {};
                \draw[red] (h2\i) -- (out2\j);
            }
        }

        \foreach \i in {1,...,5} {
            \fill [violet, opacity=0.5] (out2\i) circle (0.2);
        }

        \foreach \i in {1,...,5}{
            \node [circle,draw,inner sep=4pt, thick] (h3\i) at (3,-\i) {};
            \fill[violet, opacity=0.5] (h3\i) circle (0.2);
        }

        \foreach \j in {1,...,5} {
            \foreach \k in {1,...,2} {
                \node[circle,draw,inner sep=4pt, thick] (out3\k) at (5,-\k-1.5) {};
                \draw[orange, ->] (h3\j) -- (out3\k);
            }
        }

        \draw[->, line width=1mm] (5.5,-3) -- (6.5,-3);


        \foreach \i in {1,...,5} {
            \node[circle,draw,inner sep=4pt, thick] (h4\i) at (7,-\i) {};
        }

        \foreach \i in {1,...,5} {
            \foreach \j in {1,...,2} {
                \node[circle,draw,inner sep=4pt, thick] (out4\j) at (9,-\j-1.5) {};
                \draw[orange, ->] (h4\i) -- (out4\j);
            }
        }

        \foreach \i in {1,...,2} {
            \fill [magenta, opacity=0.5] (out4\i) circle (0.2);
        }
        
        \node[above=0.3cm] at (h1) {$k$};
        \node[above=0.3cm] at (out1) {$k+1$};
        \node[above=0.3cm] at (h21) {$k$};
        \node[above=0.3cm] at (out21) {$k+1$};

        \node[above=0.3cm] at (h31) {$k+1$};
        \node[above=0.3cm] at (out31) {$k+2$};
        \node[above=0.3cm] at (h41) {$k+1$};
        \node[above=0.3cm] at (out41) {$k+2$};
        
    \end{tikzpicture}

    \vspace{1cm}

    \begin{tikzpicture}[scale=0.6]

        \draw[thick] (-4,0) rectangle (0,4);
        \draw (-4,0) rectangle (-2.77,2.22);
        \fill [red, opacity=0.5] (-4,0) rectangle (-2.77,2.22);

        \draw[thick] (0.5,0) rectangle (1,4);
        \fill[opacity=0.5, blue] (1,4) rectangle (0.5,2.22);

        \node at (1.5,1.75) {\large$=$};

        \draw[thick] (2,0) rectangle (2.5,4);
        \fill[opacity=0.5, violet] (2,0) rectangle (2.5,2.22);

        \node at (-2,-0.5) {$A^{(k)}$};
        \node at (0.85,-0.5) {$\textbf{a}_k$};
        \node at (2.25,-0.5) {$\textbf{a}^\star_k$};

        \draw[thick] (8,0) rectangle (12,4);
        \draw (8,0) rectangle (10.86,1.14);
        \fill [orange, opacity=0.5] (8,0) rectangle (10.86,1.14);

        \draw[thick] (12.5,0) rectangle (13,4);
        \fill[opacity=0.5, violet] (12.5,4) rectangle (13,1.14);

        \node at (13.5,1.75) {\large$=$};

        \draw[thick] (14,0) rectangle (14.5,4);
        \fill[opacity=0.5, magenta] (14,0) rectangle (14.5,1.14);

        \node at (10,-0.5) {$A^{(k+1)}$};
        \node at (12.75,-0.5) {$\textbf{a}_{k+1}$};
        \node at (14.25,-0.5) {$\textbf{a}^\star_{k+1}$};

    \end{tikzpicture}

    \caption{{The linear information transfer across adjacent layers is encoded in a squared adjacency matrix.  Notice that the transformation from \(\textbf{a}_k^\star\) to \(\textbf{a}_{k+1}\) is a projection, mathematically implemented via a suitable block matrix \(\pi\), with all the entries set to \(0\) except for those referred to the upper right \(N_2 \times N_2\) square block, which are set to the identity.}}

\label{fig::linear_transfer}
\end{figure}

Focus on layer $k<\ell$ and consider the associated activity vector $\textbf{a}_{k}$ made of $N_k+N_{k+1}$ entries. The first $N_k$ elements of  $\textbf{a}_{k}$ reflect the signal that has reached the nodes on layer $k$. All other entries of $\textbf{a}_k$ are set to zero. Our ultimate goal is to transform the input $\textbf{a}_k$ into an output vector $\textbf{a}^\star_{k}$ , also of  size $N_k+N_{k+1}$,  whose last  $N_{{k+1}}$ elements display the intensities on the arrival nodes. This is achieved via the linear transformation $\textbf{a}^\star_{k}= {A}^{(k)} \textbf{a}_{k}$ where  ${ A}^{(k)}$ stands for the square $\left( N_k+N_{k+1} \right)  \times \left( N_k+N_{k+1} \right)$ adjacency matrix of the generic bipartite directed graph bridging layer $k$ to layer $k+1$. Further,  $\textbf{a}^\star_{k}$ is  processed via a suitably defined non-linear function $f\left( \cdot \right)$. {The above setting is graphically illustrated in Figure \ref{fig::linear_transfer}.}

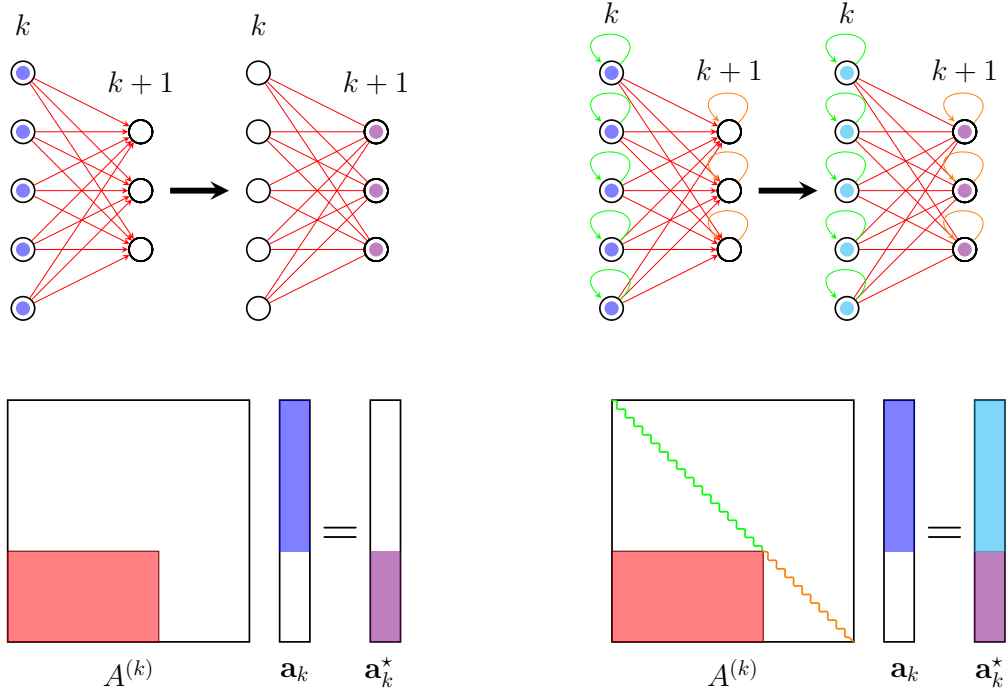
\begin{figure}[ht]
    \centering
    \resizebox{.9\textwidth}{!}{%
    \begin{tikzpicture}[>=stealth]
        \foreach \i in {1,...,5} {
            \node[circle,draw,inner sep=4pt, thick] (h\i) at (-10,-\i) {};
            \fill[blue, opacity=0.5] (h\i) circle (0.12);
        }
        
        \foreach \i in {1,...,5} {
            \foreach \j in {1,...,3} {
                \node[circle,draw,inner sep=4pt, thick] (out\j) at (-8,-\j-1) {};
                \draw[red, ->] (h\i) -- (out\j);
            }
        }
        
        \draw[->, line width=1mm] (-7.5,-3) -- (-6.5,-3);
        
        \foreach \i in {1,...,5} {
            \node[circle,draw,inner sep=4pt, thick] (h2\i) at (-6,-\i) {};
        }
        
        \foreach \i in {1,...,5} {
            \foreach \j in {1,...,3} {
                \node[circle,draw,inner sep=4pt, thick] (out2\j) at (-4,-\j-1) {};
                \draw[red] (h2\i) -- (out2\j);
            }
        }

        \foreach \i in {1,...,3} {
            \fill [violet, opacity=0.5] (out2\i) circle (0.12);
        }

        \node [above=0.5cm] at (h1) {\large$k$};
        \node [above=0.5cm] at (out1) {\large$k+1$};
        \node [above=0.5cm] at (h21) {\large$k$};
        \node [above=0.5cm] at (out21) {\large$k+1$};

        \foreach \i in {1,...,5} {
            \node[circle,draw,inner sep=4pt, thick] (h\i) at (0,-\i) {};
            \fill[blue, opacity=0.5] (h\i) circle (0.12);
        }
        
        \foreach \i in {1,...,5} {
            \foreach \j in {1,...,3} {
                \node[circle,draw,inner sep=4pt, thick] (out\j) at (2,-\j-1) {};
                \draw[red, ->] (h\i) -- (out\j);
            }
        }

        \foreach \i in {1,...,5}{
            \draw[->, green] (h\i) edge [out=45,in=135,loop] (h\i);
        }
        
        \foreach \i in {1,...,3} {
            \draw[->, orange] (out\i) edge [out=45,in=135,loop] (out\i);
        }
        
        \draw[->, line width=1mm] (2.5,-3) -- (3.5,-3);
        
        \foreach \i in {1,...,5} {
            \node[circle,draw,inner sep=4pt, thick] (h2\i) at (4,-\i) {};
            \fill[cyan, opacity=0.5] (h2\i) circle (0.12);
        }
        
        \foreach \i in {1,...,5} {
            \foreach \j in {1,...,3} {
                \node[circle,draw,inner sep=4pt, thick] (out2\j) at (6,-\j-1) {};
                \draw[red] (h2\i) -- (out2\j);
            }
        }

        \foreach \i in {1,...,3} {
            \fill [violet, opacity=0.5] (out2\i) circle (0.12);
        }
        
        \foreach \i in {1,...,3} {
            \draw[->, orange] (out2\i) edge [out=45,in=135,loop] (out2\i);
        }

        \foreach \i in {1,...,5}{
            \draw[->, green] (h2\i) edge [out=45,in=135,loop] (h2\i);
        }

        \node [above=0.7cm] at (h1) {\large$k$};
        \node [above=0.7cm] at (out1) {\large$k+1$};
        \node [above=0.7cm] at (h21) {\large$k$};
        \node [above=0.7cm] at (out21) {\large$k+1$};

    \end{tikzpicture}
    }
    
    \vspace{1cm}
    
    \resizebox{.9\textwidth}{!}{%
    \begin{tikzpicture}

        \tikzset{decoration={snake,amplitude=.4mm,segment length=2mm, post length=0mm,pre length=0mm}}

        \draw[thick] (-10,0) rectangle (-6,4);
        \draw (-10,0) rectangle (-7.5,1.5);
        \fill [red, opacity=0.5] (-10,0) rectangle (-7.5,1.5);


        \draw[thick] (-5.5,0) rectangle (-5,4);
        \fill[opacity=0.5, blue] (-5,4) rectangle (-5.5,1.5);

        \node at (-4.5,1.75) {\huge$=$};


        \draw[thick] (-4,0) rectangle (-3.5,4);
        \fill[opacity=0.5, violet] (-4,0) rectangle (-3.5,1.5);

        \node at (-8,-0.5) {\large$A^{(k)}$};
        \node at (-5.25,-0.5) {\large$\textbf{a}_k$};
        \node at (-3.85,-0.5) {\large$\textbf{a}^\star_k$};
        
        \draw[thick] (0,0) rectangle (4,4);
        \draw (0,0) rectangle (2.5,1.5);
        \fill [red, opacity=0.5] (0,0) rectangle (2.5,1.5);
        \draw[decorate, green, thick] (0,4) -- (2.5,1.5);
        \draw[decorate, orange, thick] (2.5,1.5) -- (4,0);


        \draw[thick] (4.5,0) rectangle (5,4);
        \fill[opacity=0.5, blue] (5,4) rectangle (4.5,1.5);

        \node at (5.5,1.75) {\huge$=$};


        \draw[thick] (6,0) rectangle (6.5,4);
        \fill[opacity=0.5, violet] (6,0) rectangle (6.5,1.5);
        \fill[opacity=0.5, cyan] (6,1.5) rectangle (6.5,4);

        \node at (2,-0.5) {\large$A^{(k)}$};
        \node at (4.85,-0.5) {\large$\textbf{a}_k$};
        \node at (6.25,-0.5) {\large$\textbf{a}^\star_k$};
        
    \end{tikzpicture}
    }

    \caption{{The information flow for a network with self loops (on the right) and without loops (on the left) are compared. Notice that the bottom elements of \(\textbf{a}^\star_k\) are identical in both cases, yielding  exactly the same \(\textbf{a}_{k+1}\), upon  application of the projection matrix \(\pi\).}}

\label{fig::linear_transfer_with_diag}
\end{figure}

Focus on ${A}^{(k)}$. As we shall clarify in the following, and building on the usual prescriptions, ${A}^{(k)}$
has a rather specific structure: it displays a $N_{k+1} \times N_k$ non trivial block under the main diagonal. All other entries, including the diagonal elements, are identically equal to zero. At variance, the spectral parametrization amounts to posit ${A}^{(k)}={\Phi}^{(k)} {\Lambda}^{(k)} \left({\Phi}^{(k)}\right)^{-1}$, and deal with the eigenvalues and eigenvectors as the optimization target in dual space. Notably, and as opposed to what recalled above, matrix 
${A}^{(k)}$ has non trivial diagonal entries, the associated eigenvalues, which bear no immediate reflex on the relevant portion of the output vector $\textbf{a}^\star_{k}$ (the second bunch of $N_{k+1}$ entries), see Figure \ref{fig::linear_transfer_with_diag}. However, each eigenvalue act as a veritable knot to simultaneously adjust bundles of weights departing (or landing) on a given node. 

\begin{figure}[ht]
    \centering
    \begin{subfigure}[t]{0.3\textwidth}
        \centering
        \resizebox{\linewidth}{!}{%
            \input{Phi_1.tex}
        }
        \caption{Eigenvector matrix \(\Phi\). The block \(\phi\) is a generic block of real values, and the diagonal line indicates elements identically equal to one. The remaining entries are set to zero.}
        \label{fig:Phi}
    \end{subfigure}
    \hspace{0.02\textwidth}
    \begin{subfigure}[t]{0.3\textwidth}
        \centering
        \resizebox{\linewidth}{!}{%
            \input{Lambda_1.tex}
        }
        \caption{Eigenvalue matrix \(\Lambda\). The wavy diagonal line indicates elements that generally have values different from \(0\) or \(1\).}
        \label{fig:Lambda}
    \end{subfigure}
    \hspace{0.02\textwidth}
    \begin{subfigure}[t]{0.3\textwidth}
        \centering
        \resizebox{\linewidth}{!}{%
            \input{A_1.tex}
        }
        \caption{Adjacency matrix \(A\) produced by the spectral method.}
        \label{fig:A}
    \end{subfigure}
    \caption{Illustration of the matrices: (a) \(\Phi\), (b) \(\Lambda\), and (c) \(A = \Phi \Lambda \Phi^{-1}\).}
    \label{fig:matrices}
\end{figure}
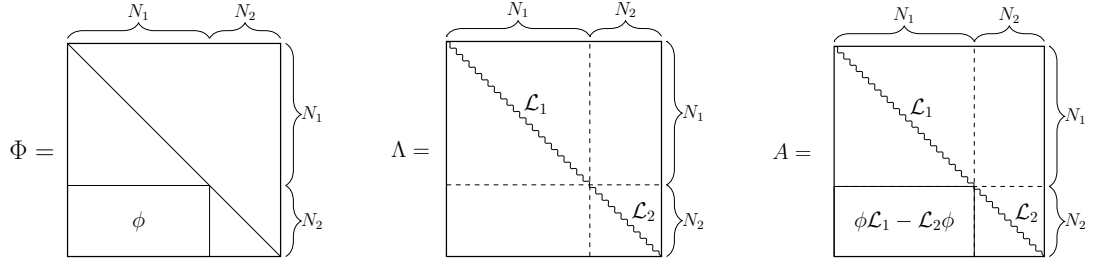

To further elaborate along these lines and to set the stage for the generalization to be subsequently discussed, we specialize on a simple perceptron-like architecture (a two layers feedforward network) by setting $\ell=2$ and denote ${A} \equiv {A}^{(1)}$. The structure of matrices $\Phi \equiv \Phi^{(1)}$, $\Lambda \equiv \Lambda^{(1)}$ and $A$ is pictorially depicted in Figure \ref{fig:matrices}. The inverse of matrix $\Phi$ can be computed analytically and reads $\left({\Phi}\right)^{-1} =  2 \mathbb{I} - {\Phi}$ where $\mathbb{I}$ denotes  the $\left( N_1+N_{2} \right)  \times \left( N_1+N_{2} \right)$ identity matrix. 

The matrix \(A\) is a bipartite directed graph where \(N_1\) nodes project onto \(N_2\) nodes through a set of weighted connections \(\mathcal{W} = \phi \mathcal{L}_1 - \mathcal{L}_2 \phi\). Each individual connection can be expressed as
\[
    {w}_{ij} = \phi_{ij} [\mathcal{L}_1]_j - [\mathcal{L}_2]_i \phi_{ij},
\]
thereby uniquely linking each component of matrix \(A\) to those of \(\Phi\) and \(\Lambda\). By spectral parametrization we intend to identify the reparametrization of the weighted connections \(\mathcal{W}\) in terms of the components of the matrices \(\Phi\) and \(\Lambda\). The term {\it spectral} is used because these components can be interpreted as the eigenvectors and eigenvalues of the adjacency matrix of the underlying graph. Rephrased differently, the spectral structure of a generic bipartite directed graph provides a novel parametrization of weights that considers the graph structure of the network.

During training, the gradient will be computed with respect to the new parameters, \(\phi, \mathcal{L}_{1,2}\), which represent the spectral degrees of freedom. This training procedure will, therefore, be named 	Spectral Learning. As remarked above, the $k-$th layer transfer across a multilayer perceptron can be represented as a bipartite network as encoded in matrix ${A}^{(k)}$. Casting ${A}^{(k)}$ via its spectral analogue allows for a straightforward generalization of the above recipe to arbitrarily deep feedforward neural networks. The output as produced on layer $\ell$ can be expressed as a chain of linear and non linear transformation as made hereafter explicit:  

\begin{equation}
\label{image}
\textbf{a}_{\ell} = \pi f\left({A}^{(\ell-1)}... \pi f\left ({A}^{(2)}  \pi f \left ({A}^{(1)} \textbf{a}_1 \right) \right) \right) 
\end{equation}
{where $\pi$ represents an appropriate projection matrix, that effectively implements the transfer of the activations from bottom to upper elements (while filling the bottom portion of the vector with zeros).}

As we shall remark in the following, the update rule for the general setting where skip layer connections are accommodated is not straightforward and comes with explicit prescriptions which rest on strategic choices. In the next Section we shall swiftly move towards discussing the novel spectral framework with reference with the simplified setting where three adjacent layers are solely accounted for. These are the input  and  output layers, interposed with one hidden layer.

\section{A simple network with just one hidden layer }

Instead of dealing with a bipartite graph, as for the case of a two-layered perceptron-like architecture, we shall here consider a directed multipartite graph. In analogy with the above, we will start from the  spectral representation, focusing in particular on the structure of the associated eigenvector matrix. To clarify the main logic path, we will refer to the three-layer setting as a simple benchmark model, before turning to consider the relevant case of a generic \(\ell\)-layer network.   

\begin{figure}[ht]
    \centering
    \begin{subfigure}[t]{0.45\textwidth}
        \centering
        \resizebox{\linewidth}{!}{%
            \input{Phi_2.tex}
        }
        \caption{Structure of a spectral matrix \(\Phi_2\) with two sub-diagonal blocks.}
        \label{fig:Phi2}
    \end{subfigure}
    \hspace{0.02\textwidth}
    \begin{subfigure}[t]{0.45\textwidth}
        \centering
        \resizebox{\linewidth}{!}{%
            \input{Lambda_2.tex}
        }
        \caption{Structure of a diagonal matrix \(\Lambda_2\) with three blocks along the diagonal.}
        \label{fig:Lambda2}
    \end{subfigure}
    \hspace{0.02\textwidth}
    \begin{subfigure}[t]{0.8\textwidth}
        \centering
        \resizebox{\linewidth}{!}{%
            \input{A_2.tex}
        }
        \caption{Structure of the adjacency matrix \(A_2\) encoded by the spectral matrix \(\Phi_2\) with two sub-diagonal blocks.}
        \label{fig:A2}
    \end{subfigure}
    \caption{Illustration of the matrices: (a) \(\Phi_2\), (b) \(\Lambda_2\), and (c) \(A_2 = \Phi_2 \Lambda_2 \Phi_2^{-1}\).}
    \label{fig:matrices2}
\end{figure}
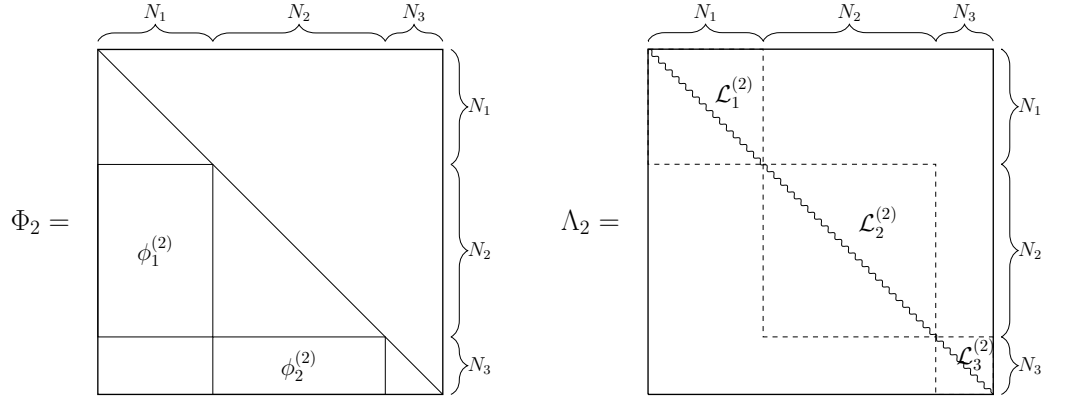
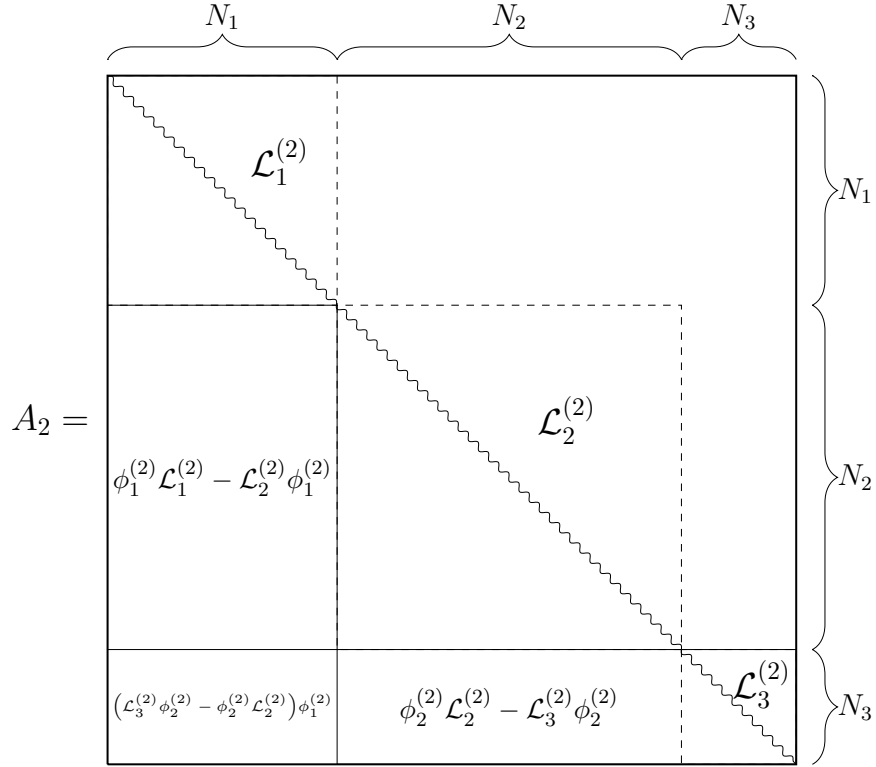

Assume an eigenvector matrix with two lower diagonal blocks that account for the presence of the hidden and output layers. The structure of such a matrix is shown in Figure \ref{fig:Phi2}. The off-diagonal blocks are denoted by \(\phi_{1,2}^{(2)}\) to emphasize that they correspond to the configuration with an input layer and two additional layers.

Remarkably, the inverse of the matrix \(\Phi_2\) can also be derived analytically to yield $\Phi_2^{-1}=\Phi_2^2-3 \Phi_2 - 3 \mathbb{I}$. The above formula is a particular case of a general theorem that we shall prove in the following when dealing with networks made of an arbitrary number of layers. Using the explicit inverse of \(\Phi_2\), the structure of the corresponding adjacency matrix \(A_2 = \Phi_2 \Lambda_2 \Phi_2^{-1}\) can be fully determined. In this case, the eigenvalue matrix \(\Lambda_2\) is divided into three different blocks associated with distinct sets of eigenvalues, one for each layer, see Figure \ref{fig:Lambda2}. The adjacency matrix \(A_2\) is shown in Figure \ref{fig:A2}.

The postulated structure of \(\Phi_2\) introduces three different sets of weights in the graph, which we have chosen to denote \(\mathcal{W}_{ij}^{(2)}\). These latter represent the connections from the \(j\)-th to the \(i\)-th layer. These weights depend on the spectral degrees of freedom, \(\phi^{(2)}_i\) and \(\mathcal{L}^{(2)}_j\), as it is specified in the following formulae:
\begin{equation}
    \mathcal{W}_{21}^{(2)} = \phi ^{(2)}_1\mathcal{L}^{(2)}_1-\mathcal{L}^{(2)}_2\phi _1^{(2)}
    \label{W21}
\end{equation}
\begin{equation}
    \mathcal{W}_{32}^{(2)} = \phi ^{(2)}_2\mathcal{L}^{(2)}_2-\mathcal{L}^{(2)}_3\phi _2^{(2)}
    \label{W32}
\end{equation}
\begin{equation}
    \mathcal{W}_{31}^{(2)}=\Big(\mathcal{L}^{(2)}_3\phi ^{(2)}_2-\phi ^{(2)}_2 \mathcal{L}^{(2)}_2\Big)\phi ^{(2)}_1
    \label{W31}
\end{equation}

It should be remarked that the spectral description of the considered multipartite graph generates a set of conventional feedforward connections, as encapsulated in the blocks \( \mathcal{W}_{21}^{(2)} \) and \( \mathcal{W}_{32}^{(2)} \). Moreover, skip connections from the first layer to the last are also established and parametrized via \( \mathcal{W}_{31}^{(2)} \). It is noteworthy that the structure of the matrix \( \Phi_2 \) naturally induces skip connections and allows for a fine control over the topology, based on the associated eigenvalues.

More specifically, the entries of \( \mathcal{L}^{(2)}_i \) control the paths made available to the network. Set, for example, \( \mathcal{L}^{(2)}_1 = \mathcal{L}^{(2)}_2 = \mathbb{O} \) and \( \mathcal{L}^{(2)}_3 = \id \).  It is immediate to appreciate that the ensuing network can only feed from the input to the output layer, thus behaving as a simple perceptron. This observation will become crucial in the following.

For now, we anticipate that by modulating the strengths of possible connections through the eigenvalues, we can continuously explore the class of underlying feedforward structures, enabling a smooth selection of the architecture. It is particularly remarkable that a consistent characterization of the adjacency matrix in terms of its associated eigenvectors and eigenvalues enables one to explore networks' inherent architectural flexibility, as we shall further explain in the following. 

Before moving on to the description of a deep feedforward network setting of arbitrary dimensions, we would like to emphasize that the postulated decomposition does not involve any diagonalization during training. Once the number of neurons is fixed, spectral training proceeds by implementing the spectral parametrization of each inter-layer bundle of connections and computing the gradient with respect to \( \phi_{1,2} \) and \( \mathcal{L}_{1,2,3} \). Moreover, it is important to note that the number of free parameters is basically the same as that of a fully connected feedforward neural network: the size of the off-diagonal blocks of matrix \( \Phi \), namely the blocks $\phi_{1,2}$, are identical to that of the weight matrices in an analogous network without skip connections. Under this parametrization, however, we gain the ability to describe different topologies, all within the same class, using fundamentally an identical number of free parameters. As a further note, it should be stressed that, under the spectral methodology, the eigenvalues which are initially set to zero can get reactivated upon training, also if a ReLU function is imposed as a non-linear filter, by virtue of their non-local character. A single eigenvalue, indeed, modulates the value of multiple connections, ensuring a non zero back-propagating gradient through the active communication channels. At variance, it is less straightforward to reactivate null silent weights in direct space, due to the vanishing of the corresponding gradient. 

Before ending this Section, we will elaborate on the different update rules that come along the representation of the network in terms of the associated adjacency matrix. The first viable possibility is that of applying matrix $A$ to an input vector $\textbf{a}_{1}$ made of $N_1+N_2+N_3$ elements. The first $N_1$ entries of $\textbf{a}_{1}$ are filled with the information provided to the input layer, while the remaining $N_2+N_3$ are identically equal to zero. Upon application of $A$ (followed by the non linear function $f(\cdot)$) yields $\textbf{a}_{2}= f\left( A \textbf{a}_{1} \right)$. The second set of $N_2$ elements of $\textbf{a}_{2}$ are now populated, as reflecting the transfer of content from the first to the second layer. The skip layer connections are responsible for the migration of the information from the first to the final layer, living an imprint in the last $N_3$ elements of $\textbf{a}_{2}$. The produced output (the activity delivered in the final layer or equivalently the signal displayed in the final $N_3$ entries of $\textbf{a}_{2}$) after one application of matrix $A$, is not sensing the linear part of the transformation that feeds from the second to the third layer. To overcome this limitation, one can apply $A$ twice, following the scheme $\textbf{a}_{3}= A f\left( A \textbf{a}_{1} \right)$, with an obvious meaning of the symbols involved. {Also notice that when employing the global activity vector (which spans nodes across the entire network), one can remove the projection matrix  \(\pi\) from the update formula for signal propagation.} A strategy which can be alternatively persecuted is to 
extract the connection blocks from the full $(N_1+N_2+N_3)\times (N_1+N_2+N_3)$ adjacency matrix $A$ and combine them in a suited update rule to link an input vector $\textbf{x} \in \mathbb{R}^{N_1}$ to the expected output $\textbf{y} \in \mathbb{R}^{N_3}$. In the following, 
we will choose to ignore all self-connections (recurring connections) by solely focusing on inter-layers bridges as stipulated by the transformation: 

\begin{equation}
        \textbf{y} =\mathcal{W}^{(2)}_{31}\textbf{x} + \mathcal{W}^{(2)}_{32} f \big( \mathcal{W}^{(2)}_{21}\textbf{x}\big)
    \end{equation}
where ${W}^{(2)}_{ij}$ with $i,j=1,2,3$ are self-consistently defined as a function of the spectral attributes of $A$. This is the setting we have chosen to operate with and that can be readily extended to encompass larger networks, as those discussed in the forthcoming Section.

\FloatBarrier
\section{Extending the spectral theory to deep networks of arbitrary size}

In this Section, we will show how the above results can be extended to a neural network of arbitrary depth. In the following, we will place the emphasis on recapitulating on the main conclusions. The proofs of the mathematical results will be relegated in dedicated Appendices.

\begin{figure}[ht]
	\centering
	\includegraphics[width=0.8\textwidth]{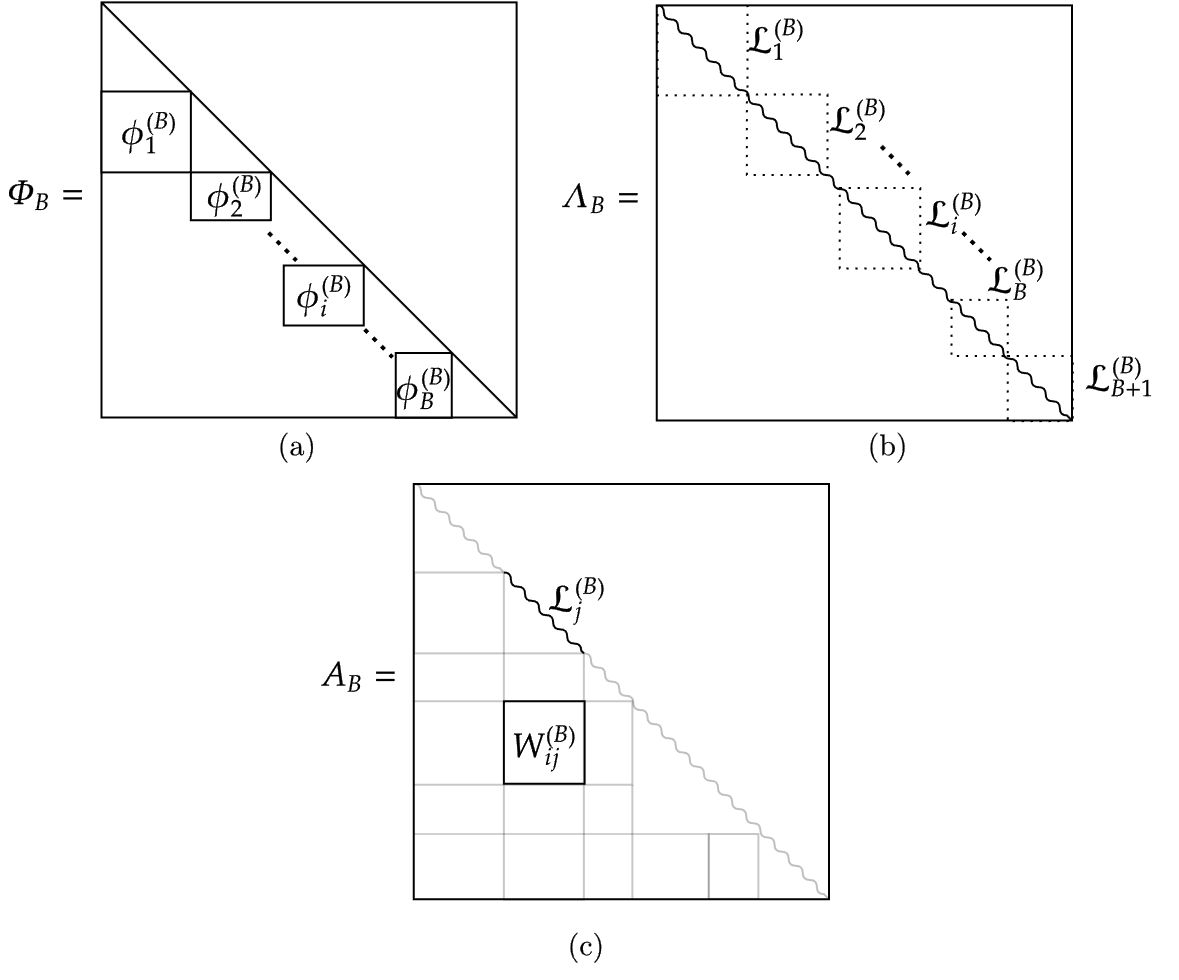}
	\caption{Schematic representation of all the matrices involved in the spectral parameterization of a generic feedforward neural network with \( B+1 \) layers. In panel (a) the eigenvector matrix, in (b) the eigenvalue matrix, and in (c) the corresponding adjacency matrix.}
	\label{fig:general_case}
\end{figure}

Assume a deep neural network with \( B+1 \) layers. Building on the above, we introduce the subdiagonal block matrix depicted in Figure \ref{fig:general_case}a as the eigenvectors matrix.  Each lower diagonal block is denoted by \( \phi_i^{(B)} \) with \( i \in 1 \cdots B \). Similarly, we define the eigenvalues' matrix \( \Lambda_B \) whose structure is pictorially represented in  panel \ref{fig:general_case}b. Since there are \( B+1 \) layers, the total number of diagonal blocks is 
 \( B \). Each block of the collection is denoted by \( \mathcal{L}_j^{(B)} \) with \( j \in 1 \cdots B+1 \). Remarkably, the matrices \( \Lambda_B \) and \( \Phi_B \) form a spectral decomposition for a lower triangular matrix whose structure is illustrated in Figure \ref{fig:general_case}c. The matrix \( \Phi_B \) is invertible, and each component of \( \Phi_B^{-1} \) can be analytically linked to the components of \( \Phi_B \) through a closed formula, which generalizes the one introduced in the preceding Section for the three layers setting. In the annexed Appendix \ref{App1} we prove that:

\begin{equation}
        \Phi _B^{-1} = \sum _{i=0}^B (-1)^i\Phi_B^i \binom{B+1}{i+1} \quad \forall B\in\mathbb{N}^{+}.\label{formula_inverse_main}
\end{equation}

Hence, each block of the ensuing adjacency matrix $A_B$  can be computed as the straightforward implementation of the product \( A_B = \Phi_B \Lambda_B \Phi_B^{-1} \). Having access to \( \Phi_B^{-1} \),  the analytical inverse for \( \Phi_B \), enables us to proceed in the analysis without relying on numerical algorithms for matrix inversion with a non negligible connected computational cost. Starting from expression (\ref{formula_inverse_main}), one can further derive explicit expressions for the subdiagonal blocks $S^{(B)}_{i,j}$ of which \(\Phi_B^{-1}\) is composed. Following a cumbersome derivation that is entirely reported in Appendix \ref{App2} we eventually obtain:

\begin{equation}
	   S^{(B)}_{ij} =
	   \begin{aligned}
		\begin{cases}
			\mathbb{O} \ \ & \text{if} \ \ j > i \\
			\mathbb{I} \ \ & \text{if} \ \ j=i \\
			(-1)^{i-j}\prod _{k=1}^{i-j} \phi _{i-k}^{(B)} \ \ \ \ \ & \text{se} \ \ j<i
		\end{cases}
	\end{aligned} \ \ \ \forall \ B
	\label{preRelPhiInv}
    \end{equation}

Building on the above, it is hence possible to compute the explicit form of the blocks that compose matrix $A$. A somehow lengthy calculation detailed in Appendix \ref{App3} yields:

{
\begin{equation}
	\mathcal{W}^{(B)}_{i,j} = \begin{dcases}
 \phi_{i-1}^{(B)}\mathcal{L}_{i-1}^{(B)} - \mathcal{L}_{i}^{(B)}\phi_{i-1}^{(B)} &\text{for} \ i-j = 1\\
 (-1)^{i-1-j}\Big[\phi_{i-1}^{(B)}\mathcal{L}_{i-1}^{(B)} - \mathcal{L}_{i}^{(B)}\phi_{i-1}^{(B)}\Big]\prod_{k=1}^{i-1-j}\phi_{i-1-k}^{(B)} &\text{for} \ i-j > 1
 \end{dcases}
	\label{blocks_general_case}
\end{equation}
}
It is worth noticing that the generic feedforward network adjacency matrix of a \( B+1 \) layer neural network, as introduced above, belongs to the complement in \( \text{GL}(N, \mathbb{R}) \) of the Parabolic Subgroup, with \( N = \sum_{i=1}^{B+1} N_i \), and the further inclusion of non trivial diagonal. Despite being widely studied, to the best of our knowledge, no attention has been so far devoted  to the spectral properties of such a family of (modified) matrices.\\

From visual inspection of matrix \( A_B \), it is immediate to 
conclude that every possible bundle of skip connections is actively represented, as an obvious byproduct of the imposed spectral parametrization. Each layer projects in fact onto every following layer. Nevertheless, it is worth stressing that the number of degrees of freedom needed to parameterize the deep neural network supplemented with skip layers connections via the spectral methodology scales as the number of parameters employed to model the corresponding feedforward version.  
The intertwining between different parameters is dictated by the spectral decomposition and the analytical form of the inverse of the matrix \( \Phi_B \). To rephrase, for the assembled network,  a parameter-sharing mechanism  exists between different bundles of connections that span both the usual feedforward connections and the skip layers. 

When operating in the general setting (i.e. with skip connections formally accommodated for), we can initialize the system with a dedicated choice of the elements that compose the diagonal blocks $\mathcal{L}_i^{(B)}$
in such a way that it is functionalized as a veritable perceptron. 

\begin{figure}[ht]
	\centering
	\includegraphics[width=0.92\textwidth]{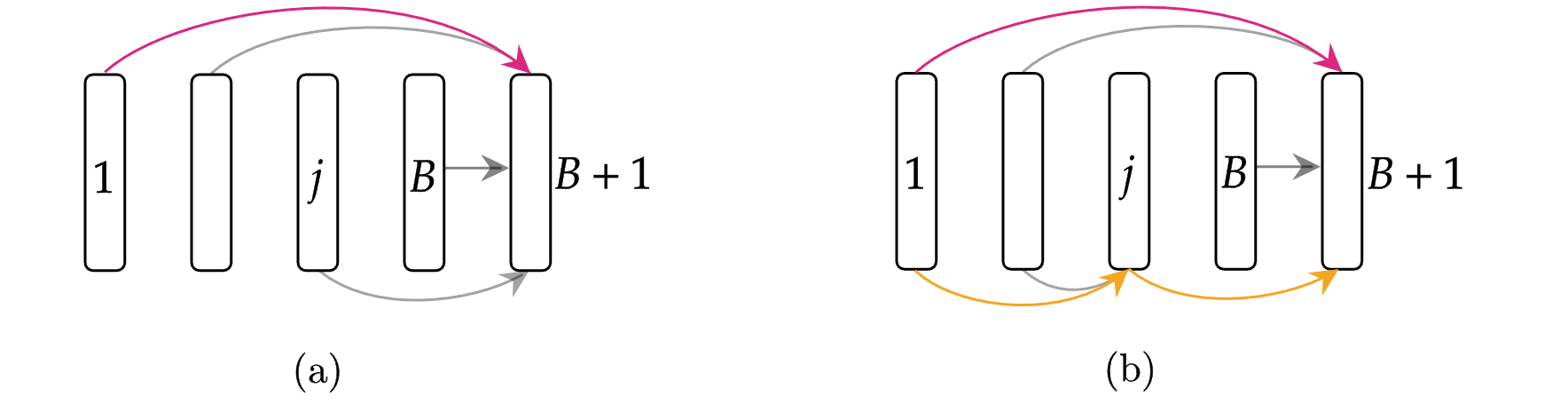}
	\caption{Panel (a): the initial configuration, starting with $\mathcal{L}_i^{(B)}=\mathbb{O}, \forall i \in 1\cdots B$ and $\mathcal{L}_{B+1}^{(B)}=\id$. Connections depicted in gray are non-zero but inactive via gradient propagation, as they do not connect with the input layer. Connections depicted in magenta represent the active bundle. Thus, the system is initialized to behave as a perceptron. In panel (b), the effect of $\mathcal{L}_j \neq \mathbb{O}$ is shown. A new significant path is formed, depicted in orange, which connects the input layer to layer $j$ and eventually to the output. Another bundle connecting every layer before $j$ to $j$ is also formed but with no material impact, as it is disconnected from the input layer when the information to be propagated is presented for further processing.}
	\label{fig:Layer_activation}
\end{figure}

Specifically, if $\mathcal{L}_i^{(B)}=\mathbb{O}, \forall i \in 1\cdots B$ and $\mathcal{L}_{B+1}^{(B)}=\id$, the only active matrices are  $\mathcal{W}^{(B)}_{B+1,j}$,  according to what stipulated by Eq. \eqref{blocks_general_case}. Every transfer between layers is hence silenced, except for those that are heading towards the last layer. Recall however that the supplied input is provided at the entry layer. Thus the   
sole connections that get \textit{de facto} operated are those bridging the gap from the input to the output layer, where the results of the analysis are eventually displayed. In this respect, the neural network calibrated as the above reacts to an external stimuli as a factual perceptron, for what it concerns the linear update rule. The described situation is graphically illustrated in Figure \ref{fig:Layer_activation} (a), where existing (grey) and active (magenta) connections are respectively depicted. 

In principle, upon training, each connection could turn active, to dynamically alter the network topology. This is due to the influence of the gradient, which is affected by parameter sharing. 
To further elaborate on this aspect, we need to consider two key points: (i) Activating $\mathcal{L}_k^{(B)}$ affects the connections coming from every layer $j<k$  due to \eqref{blocks_general_case}.
(ii) The gradient with respect to $\mathcal{L}_k^{(B)}$ will be non zero, due to matrices $\phi_B$ and $\Phi_B^{-1}$.
The effect of such activation is presented in panel (b) of Figure \ref{fig:Layer_activation}, for an hypothetical setting of exclusive pedagogical value. A new bundle of connections is activated between layer $1$ and $B+1$ via layer $j$, which impacts onto the generated output, provided the networks is operated with non linear modality (multiple applications of \( A_B \), interposed by the application of suitable non linear filters). Another interesting effect is the implicit regularization that occurs by initializing the eigenvalues to zero. This latter induces a gradual increase in rank and progressively allows to explore more complex solutions \cite{abbe2023transformers}. Remarkably, as we will demonstrate in the experiments section, this approach also prioritizes solutions with smaller depths without requiring explicit regularization targeted to that aspect.
The spectral parametrization materializes therefore in a natural strategy to modulate the functional complexity of the computing network (expressed in terms of non linear elaboration) that is differentiable and trainable. Motivated by the above, we can introduce an apt regularization factor designed so as to enhance the non linear elaboration power of the computing network, only when mandatorily needed. Relevant non linear transformations will be self-consistently enforced during training stages,
depending on the inherent intricacies of the task being explored. 

Before proceeding we elaborate further on this point. SPARCS enables the model to be initialized  with the minimal possible network topology: this is accomplished by initially setting all eigenvalues to zero, except for those associated to the neurons that belongs to the output layer. During training, the optimization of the spectral parameters modulates existing weights while, at the same time, allowing for a self-consistent growth of the network structure. To favor convergence towards compact architectures, so as to minimize the associated computational cost during deployment, one can introduce a regularization factor which acts on the eigenvalues. More specifically, the eigenvalues associated with the hidden layers are subjected to a (L1 or L2) regularization. The eigenvalues referred to the output layer are instead left untouched, while those pertaining to the input layer are set to zero. In formulae we will consistently adopt the following training loss:
\begin{equation}
\mathcal{L} (\bm{\lambda}, \bm{\phi}, \mathcal{D}) = \mathcal{L} _{\text{data}} (\bm{\lambda}, \bm{\phi}, \mathcal{D}) + \rho \Omega (\bm{\tilde{\lambda}})
\end{equation}
where \(\bm{\lambda}, \bm{\phi}\) respectively stands for  the collection of eigenvalues and eigenvectors, and \(\mathcal{D}\) identifies the elements of the training set. The loss is hence given  by the sum of a standard loss (e.g.  MSE loss, or categorical cross-entropy loss, depending on the task at hand) and a regularization term \(\Omega\) that depends exclusively on the eigenvalues associated to the hidden neurons \(\bm{\tilde{\lambda}}\). Here, \(\rho\) is a hyperparameter that sets the strength of the imposed regularization. As we shall discuss in the following, this simple strategy reliably yields minimal network topologies across a wide range of learning tasks.
This observation paves the way to a veritable Architecture Search strategy as we shall outline in the forthcoming Section with reference to simple test case models. Further experiments will be discussed in Appendix \ref{AppCIFAR}.

Before turning to discussing the aforementioned application, we devote the remaining part of this Section to concisely elaborate on the update rule of the formulated spectral network, by generalizing beyond the setting which applies to the case with just three mutually entangled layers.  
The idea is to recursively apply matrix $A_B$ to transform the input vector $\textbf{a}_{1}$ into the produced output $\textbf{a}_{B+1}$ displayed at the exit layer. Each application of $A_B$ will be followed by the implementation of the non linear function $f (\cdot)$. In formulae 
$\textbf{a}_{B+1}= A_B ... f\left( A_B f\left( A_B \textbf{a}_{1} \right) \right)$. Alternatively, one can extract the
connection blocks from the full $(\sum_{i=1}^{B+1} N_i)\times (\sum_{i=1}^{B+1} N_i)$ adjacency matrix $A_B$ and combine them to bridge vector $\textbf{x} \in \mathbb{R}^{N_1}$ to the output vector $\textbf{y} \in \mathbb{R}^{N_{B+1}}$. More precisely, and building on the three layers setting, we denote by $\textbf{a}_i$ the activation on layer $i$, with $\textbf{a}_1 \equiv \textbf{x}$, $\textbf{a}_{B+1} \equiv \textbf{y}$ and posit:

\begin{equation}
\textbf{a}_i = f \left( \sum_{k<i} \mathcal{W}^{(B)}_{ik} \textbf{a}_k \right)
\end{equation}

where ${\mathcal{W}}^{(B)}_{ik}$ with $i=1,2,..,B+1$ are functions of the spectral attributes of $A_B$. We shall set \(f\) to the identity function when computing the activation transfer to layer \(B+1\), thus effectively removing the last non-linearity. This latter is the setting that we shall adopt in the following.

\FloatBarrier
\section{Spectral Architecture Search: simple demonstrative examples}

To demonstrate the possibility of performing Neural Architecture Search via the spectral methodology, we start by considering a simple scenario in which the complexity of the function to be regressed can be tuned at will. More specifically, we introduce a family of functions dependent on two adjustable parameters, $\alpha$ and $\beta$. Parameter $\alpha$ controls a smooth transition from a simple (linear) function to a more complex (nonlinear) one, while parameter $\beta$ determines the smoothness or, conversely, abruptness of the transition. In practical term we assume:

\begin{equation}\label{family_functions}
	f(\textbf{x}) = \frac{1}{4}\left[1-\tanh\left(\beta\left(\alpha - \frac{1}{2}\right)\right)\right] \textbf{w}\cdot \textbf{x} + \frac{1}{4}\left[1+\tanh\left(\beta\left(\alpha - \frac{1}{2}\right)\right)\right]g(\textbf{x})
\end{equation}

where $\alpha \in [0,1]$ and $\beta \in (0, \infty)$. Here, $g(\textbf{x}): \mathbb{R}^d \to \mathbb{R}$ is an appropriately chosen nonlinear function.

The dataset used in this analysis is generated by sampling from the data distribution given by $p_{\text{data}}(\textbf{x}, y) = p_x(\textbf{x}) \delta(y - f(\textbf{x}))$, where $p_x(\textbf{x}) = \mathcal{U}([-1,1]^d)$ represents a uniform independent variable distribution and $d$ is the input dimension. Therefore, we are considering a deterministic supervisor.

By using the above parameterization, we can smoothly interpolate between a purely linear regression ($\alpha = 0$) and a fully nonlinear one ($\alpha = 1$). The steepness of the smooth transition is controlled by $\beta$. As $\beta \to \infty$, the transition becomes discontinuous, and take place at $\alpha = \frac{1}{2}$.

This dataset allows us to highlight the spectral parametrization's capabilities by tracking the phase transition through the number of non-zero eigenvalues associated with the intermediate neurons. In the preceding Section, we discussed how the spectral protocol introduces parameters (the eigenvalues) naturally associated with the neural network's nonlinearity. These eigenvalues account for the presence of non-zero connection paths from the input layer to the output layer through one or more intermediate layers.

When these eigenvalues, denoted as $\mathcal{L}_j$ (see Figure \ref{fig:Layer_activation}), are set to zero, the network simplifies into a perceptron, which effectively feeds from the input to the output layer without intermediate processing. This is the initialization of the network, before training takes place. A perceptron without the inclusion of non-linearity can only explain linear data. It is therefore reasonable to argue that the neural network, in its original design, will be only capable to handling data generated for $\alpha$ close to zero. As soon as $\alpha$ takes non zero values, the training needs to recruit non linear corrections and modify the initial linear setting for the data to be correctly interpolated. It could be guessed that the number of non-zero eigenvalues (as obtained upon supervised training with data extracted from the distribution $p_{\text{data}}$) should positively correlate with the magnitude of the imposed $\alpha$, and experience a phase transition, which we expected more abrupt the larger the value of $\beta$. Importantly, the training should be supplemented with an appropriate regularization that keeps at minimum the number of non zero eigenvalues introduced by the optimization.

From an operative point of view, we set initially $\mathcal{L}_i^{(B)} = \mathbb{O}$ for all $i \in 1, \cdots, B$ and $\mathcal{L}_{B+1}^{(B)} = \mathbb{I}$, a choice that, as already remarked, corresponds to operate with a perceptron. We then optimize a quadratic loss function with a weight decay applied to the parameters $\mathcal{L}_i^{(B)}$. After training, we will count with an indirect proxy the number of non-zero eigenvalues as a function of $\alpha$ for different values of $\beta$. In the reported experiments we will set $B=2$ (hence deal with a three layers network to interpolate the sought function).

\begin{figure}[ht]
	\centering
	\includegraphics[width=\textwidth]{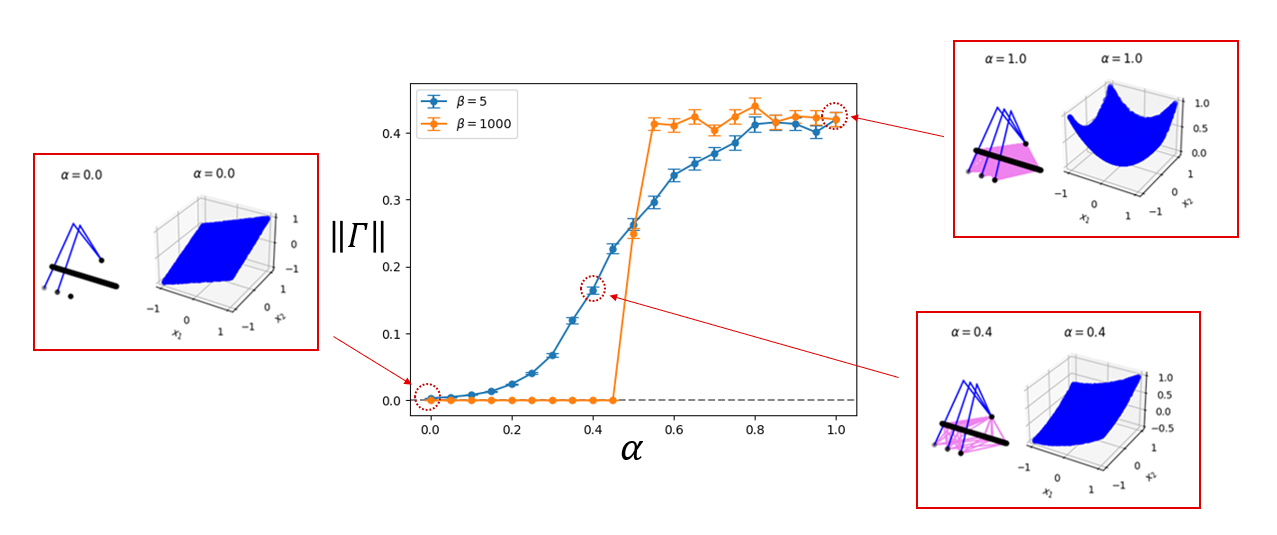}
	\caption{Main panel: the Frobenius norm of the matrix $\Gamma$ (as defined in the main text) as a function of the control parameter $\alpha$, for two different scenarios. The norm undergoes a phase transition, whose discontinuity can be tuned via the parameter $\beta$. Specifically, for $\beta = 5$, the task complexity changes smoothly with $\alpha$, and this transition is accurately reflected in $\|\Gamma\|$. Conversely, an abrupt change is observed around $\alpha = \frac{1}{2}$ for $\beta = 10^3$, as defined by the functional family in eq. \ref{family_functions}. Each annexed panel displays the final structure (upon training) of the three layer neural network, together with the function to be regressed, for a few paradigmatic choices of $\alpha$ (and with $\beta = 5$). The left side of each panel unequivocally proves that the network topology adapts to the nonlinearity requirements, thanks to the connections established via the spectral parameterization.}
	\label{fig:norm_eigval}
\end{figure}

Figure \ref{fig:norm_eigval} reports the obtained results for a specific choice of the nonlinear function $g(\textbf{x}) = \textbf{x} \cdot \textbf{x}$ and $\textbf{w} = \textbf{1}$. Form visual inspection, it appears evident that the eigenvalues belonging to the intermediate layer are indeed crucial to reproduce the nonlinear character of the target function (see annexed panels). Eigenvalues enable the network to dynamically explore the best possible topology given the supplied dataset, a non trivial ability which is encoded in the spectral representation and that we interpret as an early form of architecture search. Once training is complete, the effective topology, along with the corresponding weights, can be extracted and used as a compact model for inference.\\

We then inspect the (linear) path magnitude from the input layer to the output layer going through the intermediate one. To this end, we consider the path $i \to j \to k$, where $i\in \{1\dots N_0\}, \, j\in \{1\dots N_1\},\,k \in \{1\dots N_2\}$, namely the value of the tensor $\Gamma_{ijk} = \mathcal{W}^{(2)}_{kj}\mathcal{W}^{(1)}_{ji}$.
In Figure \ref{fig:norm_eigval} the Frobenius norm of the tensor $\Gamma$ is plotted as a function of $\alpha$, for two different choices of $\beta$: a small one, corresponding to a continuous interpolation between linear and non linear dataset, and a large one in relation with a discontinuous change at $\alpha =\frac{1}{2}$. The reported trend is in line with the expectations and testifies on the ability of the trained network to adjust to the inherent complexity (here, linear vs. non linear) of the explored dataset. Additional information on the experiments that we carried out (as well as technical details about the employed settings) are reported in Appendix \ref{info_experiments}.

\begin{figure}[ht]
    \centering
    \includegraphics[width=.6\textwidth]{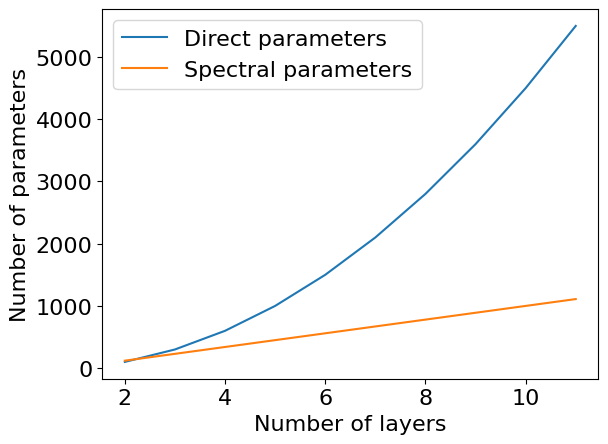}
    \caption{Number of parameters needed (assuming constant layer dimension equal to \(100\)) to perform architecture search against the total number of employable layers (not just the number of hidden layers). The scaling indicates that the spectral formalism (orange line) requires significantly fewer parameters, except for the two-layer case, where direct parameterization (blue curve) has a slight advantage.}
    \label{figure_param_adv}
\end{figure}

To fully appreciate the interest of the proposed architecture search strategy, anchored in reciprocal domain, we elaborate on the benchmark setting formulated in direct space. In this latter case, every skip connection would have been parametrized, independently from any other. Conversely, under the spectral paradigm, eigenvalues act as global parameters which can simultaneously influence a large pool of inter-nodes weights, as specified by the formulae derived above. To draw a comparison, we consider a multi-layer perceptron, with a varying number of intermediate layers, each made of an identical number of neurons (here set to 100). When varying the number of hidden layers we compare the number of free parameters that are to be adjusted for performing architecture search respectively in direct (acting on the weights) and spectral (working with the eigenvalues) settings. In \ref{figure_param_adv}, the advantage of operating with the parameter sharing induced by the spectral formalism (orange curve) is evident. The number of parameters required under the spectral framework is ultimately set by the size of the blocks of matrix $\Phi_B$, as depicted in Figure \ref{fig:general_case} (a), and by the total number of computing neurons that compose the interacting layers. 
This is approximately equivalent to the number of parameters in multi-layer perceptron with the same number of hidden layers but no skip connections (since the number of eigenvalues scales linearly with the number of neurons). In contrast, standard parametrization, without spectral adjacency matrix-induced parameter sharing, requires a number of degrees of freedom equal to the sum of the sizes of each lower diagonal block in matrix $A_B$ in Figure \ref{fig:general_case} (c).

The first test involving $f(\textbf{x})$ has clearly shown that SPARCS is capable to effectively capture the  transition in complexity of the function to be regressed, with an activation pattern which positively correlates with the displayed degree of inherent complexification. Now we will show how such response also prioritizes solutions with low architectural complexity, namely less layers.

To elaborate on this effect, we designed the following simple experiment. We construct the function $t(\textbf{x})$ as
\begin{equation}
    t(\textbf{x}) = W^{(2)}\sigma(W^{(1)}\textbf{x})
\end{equation}
with $W^{(1)}, W^{(2)} : \mathbb{R}^{20} \to \mathbb{R}^{20} $ matrices. The latter are random 20-dimensional rotations, namely sampled from $\text{SO}(20)$, ensuring an empty kernel, thus conserving information. The dataset used in this analysis is, again, generated by sampling from the data distribution given by $p_{\text{data}}(\textbf{x}, y) = p_x(\textbf{x}) \delta(y - t(\textbf{x}))$, where $p_x(\textbf{x}) = \mathcal{U}([-1,1]^d)$ represents a uniform independent variable distribution and $d$ is the input dimension. We are therefore considering a deterministic supervisor that is a random neural network with just one hidden layer.
The ReLU function is then applied as a nonlinearity, followed by another high-dimensional rotation. Overall, this network structure defines a nonlinear map between input and output that requires at least one hidden layer to be eventually reproduced. In order to check that this is indeed the case, we carried out a linear regression model and computed the corresponding $R^2$ metric, to ensure that a linear, zero hidden layer, model is not enough to reproduce the examined dataset. Having a fully deterministic conditional distribution of $y$ (the Dirac Delta) and a sample of $10^5$ data points, we indeed obtained a $R^2$ score sensibly different from one due to the strongly biased linear estimator, which is thus proved not expressive enough to grasp the essence of the analyzed dataset.

\begin{figure}[ht]
    \centering
    \includegraphics[width=\textwidth]{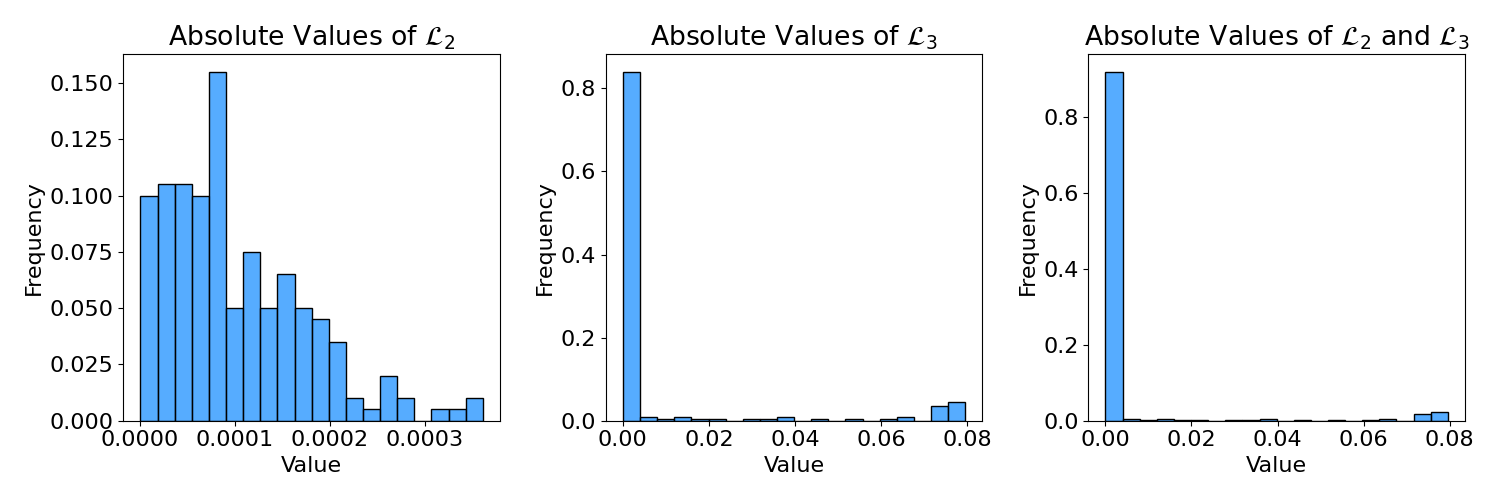}
    \caption{Histograms of the absolute values of the post-training entries of the matrices $\mathcal{L}_{2,3}$. In our proposed methodology, $\mathcal{L}_{1}$ is always set to zero, and $\mathcal{L}_{4}$ cannot be turned off; otherwise, the learning task would be impaired as output neurons would be removed. As shown, the values of $\mathcal{L}_{2}$ are much smaller than those of $\mathcal{L}_{3}$. On the right, the aggregated distribution is displayed. All eigenvalues populating the peak of the first bin of the histogram are deemed irrelevant and can thus be set to numerical zero, implementing the related architectural change.}
    \label{fig:teacher_student}
\end{figure}

A full training session is then carried out by using a maximal architecture of a two-hidden-layers fully connected feedforward network. The number of hidden neurons is set to $N_1 = N_2 = 200$, which is ten times larger than that of the neural network $t(\textbf{x})$, thus ensuring over-parameterization. Every connection is parameterized according to the SPARCS technique, with the optimization starting point characterized by a perceptron-like function and regularized eigenvalues. After training the eigenvalues can be inspected and their distribution assessed. Recall that, due to the fact that we have a total of four layers in our regressing function (two hidden), we can fix the value of $\mathcal{L}_{1,2,3,4}$. $\mathcal{L}_{4}$ is initialized to be non-zero, on the contrary $\mathcal{L}_{1}$ will be permanently fixed to zero. As a consequence the distribution to be inspected after training, which ultimately dictates the emerging structure of the regressing network, is that associated to the components $\mathcal{L}_{2}$ and $\mathcal{L}_{3}$. These latter distributions are reported on the left and central panels of Figure \ref{fig:teacher_student}. As it can be clearly appreciated by visual inspection of the reported $\mathcal{L}_{3}$ histogram, the eigenvalues separate into two distinct families: the inactive ones, populating the first peak in zero, and all the rest.
Such peculiar distribution of the trained eigenvalues can be exploited via a suitable pruning strategy named Spectral Pruning \cite{teacher_student_giambagli, SpectralPrune} whose efficacy has already been demonstrated in different scenarios. More specifically, as the eigenvalues multiply a full bundle of connection that goes into a neuron,  applying a magnitude-based pruning strategy on those parameters results in the removal of a neuron from the neural network structure. This operation is possible as nodes which bear eigenvalues orders of magnitude smaller than others have a negligible effect on the information transfer, and can hence be safely removed from the pool. Indeed, the values of $\mathcal{L}_{2}$, corresponding with second hidden layer related activations, are two order of magnitude smaller and can therefore be deemed irrelevant in the computation, thus enabling full exclusion of the first hidden layer from the architecture. 
This strategy could also be extended to the neurons of the first layer, the active one. Remarkably, indeed,  several eigenvalues (and therefore neurons) of the second hidden layer are essentially irrelevant for the ensuing computation. Those can be spotted out by plotting the relative difference of validation loss between the model with all the eigenvalues activated and the structurally pruned homologue, where the eigenvalues below a certain threshold are set to zero (which amounts in turn to remove the associated nodes from the network). 

\begin{figure}[ht]
    \centering
    \includegraphics[width=\textwidth]{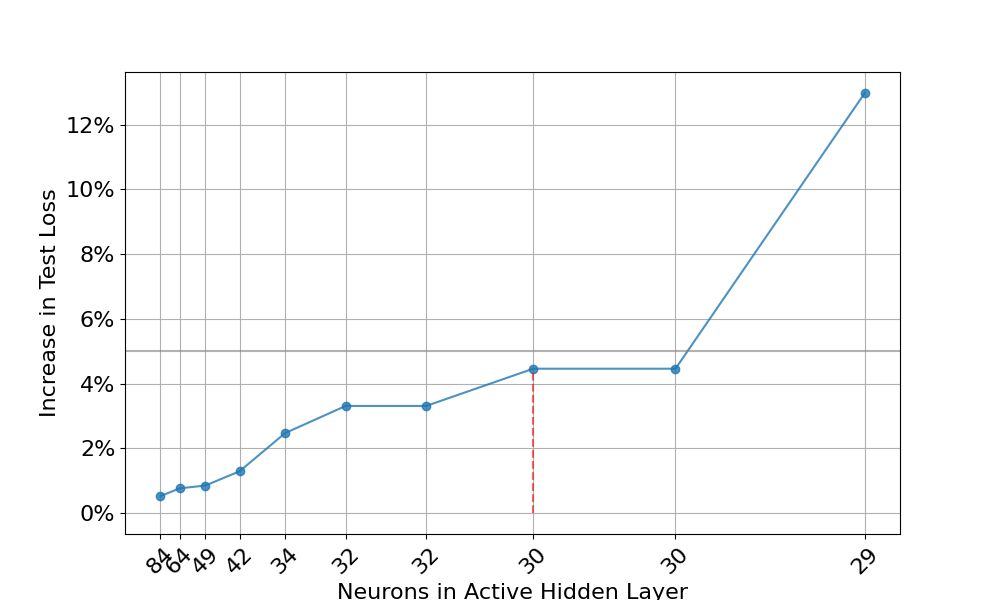}
    \caption{Curve of the relative increase in validation error due to the removal of nodes in the active hidden layer (the second one). On the $x$ axis, for simplicity, the number of active neurons is shown. With thicker gray line the (arbitrary) threshold palced at 5\%.}
    \label{fig:pruning}
\end{figure}

At the end of the procedure, we obtain the curve of Figure \ref{fig:pruning}. We arbitrarily decided to place a threshold at 5\% as the maximal increase in loss, and therefore selected the number of residual active neurons in the network, following a procedure detailed in \cite{teacher_student_giambagli}. It is worth point out that, in this scenario, the correspondence between neurons and activation is a direct consequence of equation \eqref{blocks_general_case} with $\mathcal{L}_{1}$ and $\mathcal{L}_{3}$ equal to zero.  We would like to stress that, in more general settings, the layer deactivation still holds but the node to eigenvalue correspondence might not apply on every layer.\\ Interestingly, in this controlled scenario, the model after training, having at its disposal two 200 dimensional hidden layers, ends up in just one hidden layer with a 30 neurons configuration, very similar to the one imposed by the structure of function $t(\textbf{x})$. As it can be confirmed via the available code, the $R^2$ score associated to the computed solution is large, around 0.98, at variance with what one gets with the linear model. This proves that the trimmed model provides a high fidelity fit.
Together with the first test we have shown, under simple and controlled scenarios, the capability of SPARCS to spot a solutions with a low level of complexity that results in simple architectural features both in terms of depth and width. Another significant advantage of operating under the spectral viewpoint resides in the  possibility of addressing two major issues that arise when performing architecture search with no parameter sharing. First, setting all weights in direct space to zero often halts the gradient flow through target layers, hindering the learning of complex architectures. Second, even if the gradient stays non-zero, networks initialized with symmetrical weights face challenges during training due to symmetry-breaking issues \cite{glorot2010understanding}. These problems are avoided under the spectral formalism due to the interplay with eigenvectors, which are always initialized with non-zero random values.

Two additional tests for the newly proposed algorithm are discussed in Appendix \ref{AppCIFAR} The first set of experiments are run on CIFAR-10. The architecture is composed by a minimal CNN module, that we consistently train across the experiments. The output of the CNN is then fed to a feedforward network, to perform classification. This latter feedforward network is initialized as a perceptron (a linear classifier). As in the spirit of the other reported tests, the architecture can be modulated by SPARCS and the performance gets self-consistently boosted (as compared to a benchmark linear model run in parallel) by taking full advantage of the non-linear character of the feedforward classifier. Indeed, the simple CNN module that we have employed is unable to reduce the analyzed task to a linear classification problem and SPARCS finds it advantageous to turn the classification from linear to non linear. In the second experiment, we considered CIFAR-100, a rather more complex dataset, but now we employed a pre-trained state-of-the-art convolutional neural network architecture (EfficientNetV2-S) as a pre-filter. In this case the powerful convolutional module turns the data into a  linearly separable problem and SPARCS keeps across fine-tuning the minimal linear architecture, as it follows by the imposed initialization, without recruiting additional layers of the feedforward classifiers. With reference to these applications, we report also an estimate of the involved computational costs and related memory usage. For further technical information, please refer to Appendix \ref{AppCIFAR}.

The proposed Neural Architecture Search methodology,  follows from a non trivial generalization of the spectral approach to machine learning as pioneered in \cite{spectral_learning}. For this reason, and as recalled above, it is termed  SPARCS, from SPectral ARchiteCture Search. In short, SPARCS allows one to explore the space of possible architectures by spanning continuous and differentiable manifolds, via the spectral attributes of the inter-layer transfer matrices and thus enabling for gradient-based optimization algorithms to be eventually employed.

\section{Conclusions}
    
In the first part of the work, we introduced a novel method for parameterizing the weights of a multilayer perceptron, grounded in the spectral properties of a generic multipartite graph. The proposed approach is a non trivial generalization of the {\it spectral parametrization} \cite{spectral_learning,
teacher_student_giambagli, SpectralPrune, preChicchi}, to handle the optimization of feedforward networks. More specifically,  the spectral machinery has been successfully extended beyond the setting where information is solely made to flow across two consecutive layers. 
We have in particular laid the foundations for a spectral treatment of 
the local and long-ranged interactions of an arbitrary collection of inter-tangled layers, of different sizes. This generalized framework allows for 
a wider usage of the  spectral technology, along directions of investigations that remain to be explored.

The idea of intertwining different weights within an architecture has been explored in the literature \cite{he2016deep, Denil_2013,liu2021orthogonal}. Our results indicate that, although a large number of parameters at initialization is required for speeding up training and improving generalization performance, many of these  exhibit strong correlations. These latter materializes in a consequent reduction of the truly independent degrees of freedom for the neural network under investigation. Prior research \cite{zhang2020efficient} has shown that effective training can be achieved by regressing only a smaller number of parameters, which simultaneously alter the values of a larger set of weights through a parameter-sharing mechanism. Our method can be interpreted along these lines of thought, as providing a natural parameter sharing recipe that is deeply rooted in the spectral properties of the underlying multipartite graph.

By treating a multilayer perceptron as a multipartite graph, we established an analytical correspondence between the links of any two layers (including not only adjacent inter-layer connections but also the so-called "skip-layer" connections) and the eigenvalues and eigenvectors of the graph's adjacency matrix. This parameter-sharing mechanism is induced by such an analytical correspondence as formalized in Eq. \eqref{blocks_general_case}.

Remarkably,  the proposed parameter-sharing approach enables us to initialize our network as a perceptron, with only the connections between the input and output layers being active. During training, more advanced nonlinear topologies involving hidden layers can be dynamically activated and explored as needed, without a significant increase in the number of free parameters, when confronted to a multilayer perceptron with no skip connections. The possibility of conduction Neural Architecure Search via the spectral decomposition constitutes the second contribution of this paper. Working with simple testbed models,  we have proven that the proposed framework, termed SPARCS after SPectral ARchiteCture Search, can recruit  hidden non linear neurons as needed, tailoring their activation as a function of the inherent complexity of the supplied data.
As a relevant complement, the number of free parameters that are to be adjusted thanks to the enforced spectral correlations is significantly smaller than that required in direct space. The success of SPARCS, as reported in our  preliminary applications, and the overall flexibility of the proposed mathematical framework, will motivate further research to uncover potential assets of the spectral approach to neural network modeling and design.\\
{\bf Code availability} The code employed for the tests reported in this paper is made available at the following link \url{https://github.com/gianluca-peri/sparcs-test}.\\
{\bf Acknowledgments} This work is supported by \textsc{nextgenerationeu (ngeu)} and funded by the Ministry of University and Research (\textsc{mur}), National Recovery and Resilience Plan (\textsc{nrrp}), project \textsc{mnesys} (\textsc{pe0000006}) A multiscale integrated approach to the study of the nervous system in health and disease" (\textsc{dr. 1553 11.10.2022}).

\FloatBarrier

    \newpage
    \appendix
    {
    \section{The analytic inverse of the eigenvectors' matrix \texorpdfstring{\(\Phi_B\)}{}}
     \label{App1}}

    Consider a generic \(\Phi_B\) matrix, with \(B\) concatenated sub-diagonal blocks of arbitrary dimension, as introduced in the main body of the paper. It is possible to show that \(\Phi_B^{-1}\) is given by the following expression:

    \begin{equation}
        \Phi _B^{-1} = \sum _{i=0}^B (-1)^i\Phi_B^i \binom{B+1}{i+1} \quad \forall B\in\mathbb{N}^{+}.\label{formula_inverse}
    \end{equation}
    
    To prove this statement, we only need to show that
    
    \begin{equation}
        \Phi _B \sum _{i=0}^B (-1)^i\Phi_B^i \binom{B+1}{i+1} = \mathbb{I} \ \Rightarrow \ (\Phi _B - \mathbb{I})^{B+1} = \mathbb{O} 
    \end{equation}
    
    This latter equality can be easily proven by induction on \(B\). Consider the case \(B=1\). The diagonal of \(\Phi_B\) is removed by the subtraction of \(\mathbb{I}\), and we are left with a nilpotent matrix. The inductive step can be performed by noticing that:
    \begin{equation}
		\Phi _{B+1} = \begin{pmatrix}
			\Phi _B &  \mathbb{O} \\
			\alpha & \mathbb{I}
		\end{pmatrix}
	\end{equation}
    where \(\alpha\) is an appropriate rectangular matrix. This implies:
    \begin{equation}
		(\Phi _{B+1} - \mathbb{I})^{B+1} = \begin{pmatrix}
			(\Phi _B - \mathbb{I})^{B+1} &  \mathbb{O} \\
			\alpha(\Phi _B - \mathbb{I})^{B} & \mathbb{O}
		\end{pmatrix}=\begin{pmatrix}
			\mathbb{O} &  \mathbb{O} \\
			\alpha(\Phi _B - \mathbb{I})^{B} & \mathbb{O}
		\end{pmatrix}
	\end{equation}
    where the last equality holds thanks to the inductive hypothesis. What we obtain is once again a nilpotent matrix such that:
    \begin{equation}
        (\Phi _{B+1} - \mathbb{I})^{B+2}=\mathbb{O} \quad \forall B
    \end{equation}
    The former observation ends the proof of (\ref{formula_inverse}).

    \section{Explicit expressions for the blocks of \texorpdfstring{\(\Phi_B^{-1}\)}{the inverse of phi}}
    \label{App2}

    It is important to notice that \(\Phi_B^{-1}\) will display a block structure, as follows from the block structure of \(\Phi _B\). In the following, we will denote by \(T_{i,j}^{(B)}\) the blocks of \(\Phi_B\), and by \(S_{i,j}^{(B)}\) the blocks of \(\Phi_B^{-1}\). The structure of \(T_{i,j}^{(B)}\) follows the definition of \(\Phi _B\):

    \begin{equation}
		T^{(B)}_{i,j} = \begin{aligned}\begin{cases}
			\mathbb{O} \quad & \text{if} \quad j > i \\
			\mathbb{I} \quad & \text{if} \quad j=i \\
			\phi _{j}^{(B)} \quad & \text{if} \quad j=i-1 \\
			\mathbb{O} \quad & \text{if} \quad j < i-1
		\end{cases}\end{aligned} \quad \forall \ B.
		\label{nu1}
	\end{equation}
    
    Our goal is to derive the formula for the blocks \(S_{i,j}^{(B)}\).
    To this end we can leverage on (\ref{formula_inverse}), but only if we can determine beforehand the analytical expression of the blocks of \(\Phi_B^n \forall n \in \mathbb{N}^{+}\). We shall refer to these latter blocks as \(T^{(B,n)}_{i,j}\). It is indeed possible to show that:
    \begin{equation}
		T^{(B,n)}_{i,j}=\begin{aligned}\begin{cases}
			\mathbb{O} \ \ & \text{if} \ \ j > i \\
			\mathbb{I} \ \ & \text{if} \ \ j=i \\
			\tau(i-j,n)\prod _{k=1}^{i-j} \phi _{i-k}^{(B)} \ \ \ \ \ & \text{if} \ \ j<i
		\end{cases}\end{aligned} \ \ \ \ \ \forall \ B
		\label{lemma}
	\end{equation}
	with:
	\begin{equation}
		\tau(i-j,n) \ \dot= \ \begin{aligned}\begin{cases}
			0 \ \ & \text{if} \ \ n<i-j \\
			\binom{n}{i-j} \ \ & \text{if} \ \ n \geq i-j
		\end{cases}\end{aligned}
	\end{equation}

    In the following we shall set to prove equation (\ref{lemma}). Once again we will apply a recursive induction strategy. Consider the case \(n=1\). By definition we get (\ref{nu1}) for all \(B\), and we can immediately verify that this conforms to (\ref{lemma}) for all \(B\). The next step in the inductive path consist in assuming that (\ref{lemma}) holds for \(n\) and make use of this assumption to prove the validity of (\ref{lemma}) for \(n+1\) (once again for all \(B\)). Let us begin by noticing that:
    \begin{equation}
		T^{(B,n+1)}_{i,j} = \sum _{l=1}^{B+1} T^{(B,n)}_{i,l}T_{l,j}^{(B,1)}
		\label{PrSo}
	\end{equation}
    Given (\ref{nu1}) the above expression reduces to
    \begin{equation}
		T^{(B,n+1)}_{i,j} =  T^{(B,n)}_{i,j}\mathbb{I} + T^{(B,n)}_{i,j+1}\phi _{j}^{(B)}
		\label{rif}
	\end{equation}
    From this latter relation, the case \(j > i\) and \(j=i\) follow immediately. The case \(j>i\) requires some additional calculations.  We get:
    \begin{equation}
		T^{(B,n+1)}_{i,j} =  \tau(i-j,n)\prod _{k=1}^{i-j} \phi _{i-k}^{(B)}\mathbb{I} + \tau(i-j-1,n)\prod _{k=1}^{i-j-1} \phi _{i-k}^{(B)}\phi _{j}^{(B)}
	\end{equation}
    For \(n+1<i-j\), we obtain \(\mathbb{O}\), as desired. Conversely, if \(n+1 \geq i-j\) yields:
    \begin{equation}
		T^{(B,n+1)}_{i,j} =  \left[\binom{n}{i-j} + \binom{n}{i-j-1}\right]\prod _{k=1}^{i-j} \phi _{i-k}^{(B)}.
		\label{rif2}
	\end{equation}
    Given the recurrence relation of the binomial coefficients
    \begin{equation}
        \binom{n}{k}=\binom{n-1}{k-1} + \binom{n-1}{k}
    \end{equation}
    we have:
    \begin{equation}
		T^{(B,n+1)}_{i,j} =  \binom{n+1}{i-j}\prod _{k=1}^{i-j} \phi _{i-k}^{(B)}
	\end{equation}
    that is exactly what we aimed at recovering. This concludes the proof of (\ref{lemma}).

    Given (\ref{lemma}) we can use (\ref{formula_inverse}) to get the structure of the blocks \(S_{i,j}^{(B)}\):
    \begin{equation}
	    S^{(B)}_{i,j} = \sum _{k=0}^B (-1)^k T^{(B,k)}_{ij} \binom{B+1}{k+1}
        \label{base}
    \end{equation}
    From this latter equation it is possible to show that
    \begin{equation}
	   S^{(B)}_{ij} =
	   \begin{aligned}
		\begin{cases}
			\mathbb{O} \ \ & \text{if} \ \ j > i \\
			\mathbb{I} \ \ & \text{if} \ \ j=i \\
			(-1)^{i-j}\prod _{k=1}^{i-j} \phi _{i-k}^{(B)} \ \ \ \ \ & \text{if} \ \ j<i
		\end{cases}
	\end{aligned} \ \ \ \forall \ B
	\label{RelPhiInv}
    \end{equation}
    Once again, we have to dig into the proof, by starting from (\ref{base}). We can see from (\ref{lemma}) that for \(j>i\) all terms in the summation are identically equal to \(\mathbb{O}\). 
    For \(j=i\) we get instead:
    \begin{equation}
	   S^{(B)}_{ii} = \mathbb{I} \sum _{k=0}^B (-1)^k \binom{B+1}{k+1}=\mathbb{I}
	   \label{BP}
    \end{equation}
    where the last equality holds since (see Appendix \ref{proof_binom_eqs}):
    \begin{equation}
		\sum _{k=0}^B (-1)^k \binom{B+1}{k+1} = 1 \ \ \ \forall \ B.
        \label{first_for_app_b}
	\end{equation}
    To complete the proof we have to consider the case \(j<i\) for which we get
    \begin{equation}
	   S^{(B)}_{ij} = \sum _{k=0}^B (-1)^k \tau(i-j,k)\prod _{l=1}^{i-j} \phi _{i-l}^{(B)} \binom{B+1}{k+1}
    \end{equation}
    and given the form of \(\tau(i-j,k)\):
    \begin{equation}
	   S^{(B)}_{ij} = \sum _{k=i-j}^B (-1)^k \binom{k}{i-j}\prod _{l=1}^{i-j} \phi _{i-l}^{(B)} \binom{B+1}{k+1}= \left[\sum _{k=i-j}^B (-1)^k \binom{k}{i-j} \binom{B+1}{k+1}\right]\left[\prod _{l=1}^{i-j} \phi _{i-l}^{(B)}\right].
    \end{equation}
    To conclude the proof of (\ref{RelPhiInv}) one needs to show that
    
    \begin{equation}
	\sum _{k=i-j}^B (-1)^k \binom{k}{i-j} \binom{B+1}{k+1} = (-1)^{i-j} \ \ \ \forall \ B \in \mathbb{N}^+ \ \ \ \forall \ i-j \ \Big| \ 1 \leq i-j \leq B
	\label{UD}
    \end{equation}

    Proving this last equality is not as easy as one may expect: we will start by defining \(\rho \ \dot= \ i-j\) for convenience:
    \begin{equation}
	   \sum _{k=\rho}^B (-1)^k \binom{k}{\rho} \binom{B+1}{k+1} = (-1)^{\rho} \ \ \ \forall \ B \in \mathbb{N}^+ \ \ \ \forall \ \rho \ \big| \ 1 \leq \rho \leq B
	   \label{UD2}
    \end{equation}
    and once again carry out the proof by induction on \(B\). Consider the base (or initial) case: the only possible value for \(\rho\) when \(B=1\) is \(\rho=1\). We can thus proceed by direct calculation, as follows:
    \begin{equation}
	   \sum _{k=1}^1 (-1)^k \binom{k}{1} \binom{2}{k+1} = (-1)^1 \binom{1}{1} \binom{2}{2} = -1.
    \end{equation}
    Net we perform the inductive step: assume the validity of the sought relation for \(B\) and prove it for the case \(B+1\) for all the possible \(\rho\). We shall start by noting that:
    \begin{equation}
	   \begin{aligned}
		  & \sum _{k=\rho}^{B+1} (-1)^k \binom{k}{\rho} \binom{B+2}{k+1} = \sum _{k=\rho}^{B} (-1)^k \binom{k}{\rho} \binom{B+2}{k+1} + (-1)^{B+1} \binom{B+1}{\rho} \binom{B+2}{B+2} = \\
		  & = \sum _{k=\rho}^{B} (-1)^k \binom{k}{\rho} \binom{B+2}{k+1} + (-1)^{B+1} \binom{B+1}{\rho}.
	\end{aligned}
    \end{equation}
    We can now deploy the recurrence relation for the binomial coefficients to write the last expression as:
    \begin{equation}
	   \sum _{k=\rho}^{B} (-1)^k \binom{k}{\rho} \binom{B+1}{k+1} + \sum _{k=\rho}^{B} (-1)^k \binom{k}{\rho} \binom{B+1}{k} + (-1)^{B+1} \binom{B+1}{\rho},
    \end{equation}
    Then, we make use of the inductive hypothesis to rewrite this expression as
    \begin{equation}
	   (-1)^{\rho} + \sum _{k=\rho}^{B} (-1)^k \binom{k}{\rho} \binom{B+1}{k} + (-1)^{B+1} \binom{B+1}{\rho}.
    \end{equation}
    And observing  that the last term can be included in the summation yields:
    \begin{equation}
	   (-1)^{\rho} + \sum _{k=\rho}^{B+1} (-1)^k \binom{k}{\rho} \binom{B+1}{k}.
    \end{equation}
    So, to prove (\ref{UD}), we end up needing to prove that
    \begin{equation}
		\sum _{k=\rho}^{n} (-1)^k \binom{k}{\rho} \binom{n}{k} = 0 \ \ \ \forall n,\rho
		\label{second_for_app_b}
	\end{equation}
    and this is done, once again, in Appendix \ref{proof_binom_eqs}. This concludes our proof of (\ref{RelPhiInv}).

\section{On the block structure of \texorpdfstring{\(A_B\)}{the adjacency matrix}}
\label{App3}

Having proven (\ref{RelPhiInv}) we can now deal with the block structure of \(A_B\). Following the usual notation, we shall denote the blocks of \(A_B\) as \(A_{i,j}^{(B)}\). Since by definition \(A_B=\Phi _B \Lambda _B \Phi _B^{-1}\) these blocks are given by:
    \begin{equation}
        A_{ij}^{(B)} = \sum _{k=1} ^{B+1} T _{ik}^{(B)} \mathcal{L} _{k}^{(B)} S _{kj}^{(B)}
        \label{s1}
        \end{equation}
    Due to the structure of \(\Phi_B\) the latter equation reduces to
    \begin{equation}
		A_{ij}^{(B)} =\mathbb{I} \mathcal{L}_i^{(B)} S _{i,j}^{(B)} + \phi _{i-1}^{(B)} \mathcal{L}_{i-1}^{(B)} S _{i-1,j}^{(B)}.
		\label{middleancora}
	\end{equation}
    To make progress we must now recall the structure of \(\Phi _B^{-1}\), as reported in (\ref{RelPhiInv}) to eventually get:
    \begin{equation}
        A_{i,j}^{(B)}=
        \begin{cases}
            \mathbb{O} \quad j>i\\
            \mathcal{L}_i \quad j=i
        \end{cases} 
    \end{equation}
    In reality, we are just interested in the case \(j<i\). From (\ref{middleancora}) we get:
    \begin{equation}\begin{aligned}
	   A_{ij} & = \mathcal{L}_{i}^{(B)} (-1)^{i-j}\prod _{k=1}^{i-j} \phi _{i-k}^{(B)} + \phi _{i-1}^{(B)} \mathcal{L}_{i-1}^{(B)} (-1)^{i-1-j}\prod _{k=1}^{i-1-j} \phi _{i-1-k}^{(B)} = \\
	   & = -\mathcal{L}_{i}^{(B)} \phi_{i-1}^{(B)}(-1)^{i-j-1} \prod _{k=2}^{i-j} \phi _{i-k}^{(B)} + \phi _{i-1}^{(B)} \mathcal{L}_{i-1}^{(B)} (-1)^{i-1-j}\prod _{k=1}^{i-1-j} \phi _{i-1-k}^{(B)}
    \end{aligned}\end{equation}
    Recall that in the main body of this paper we choose to rename the blocks \(A_{i,j}^{(B)}\) (with \(j<i\)) with the symbol \(\mathcal{W}_{i,j}^{(B)}\). Hence, by renaming accordingly the target variable and performing the change of index \(k \to k-1\), one obtains:
    \begin{equation}
	   \mathcal{W}_{ij}^{(B)} = -\mathcal{L}_{i}^{(B)} \phi_{i-1}^{(B)}(-1)^{i-j-1} \prod _{k=1}^{i-j-1} \phi _{i-1-k}^{(B)} + \phi _{i-1}^{(B)} \mathcal{L}_{i-1}^{(B)} (-1)^{i-1-j}\prod _{k=1}^{i-1-j} \phi _{i-1-k}^{(B)}
    \end{equation}
    Grouping the common factor one gets at last:
    \begin{equation}
	   \mathcal{W}_{ij}^{(B)} = (-1)^{i-1-j}\Big[\phi _{i-1}^{(B)}\mathcal{L}_{i-1}^{(B)}-\mathcal{L}_{i}^{(B)}\phi_{i-1}^{(B)}\Big]\prod _{k=1}^{i-1-j}\phi _{i-1-k}^{(B)}
    \end{equation}
    and this ends the proof. Notice in fact that for \(i-j=1\) the last factor of this latter expression is not present.  Equation (\ref{blocks_general_case}) is thus formally recovered.

    \section{Proof of binomial formulae}
    \label{proof_binom_eqs}

    In this Appendix we provide a proof for the employed expressions (\ref{first_for_app_b}), (\ref{second_for_app_b}). Starting with the first one, we have to prove that
	\begin{equation}
		\sum _{k=0}^B (-1)^k \binom{B+1}{k+1} = 1 \ \ \ \forall \ B.
	\end{equation}
    In the following we shall make use of a renowned property of the binomial coefficients, namely:
    \begin{equation}
		\sum _{k=0}^n (-1)^k \binom{n}{k} = 0 \ \ \ \forall \ n \in \mathbb{N}^{+}.
		\label{PII}
	\end{equation}
    Specifically, consider first of all the case of \(B\) even, and write
    \begin{equation}
	\begin{aligned}
		& \sum _{k=0}^B (-1)^k \binom{B+1}{k+1} = \sum _{k=0}^{B-1} (-1)^k \binom{B+1}{k+1} + (-1)^B \binom{B+1}{B+1} = \\
		& =\sum _{k=0}^{B-1} (-1)^k \binom{B+1}{k+1} + 1.
	\end{aligned}
	\end{equation}
    Also, by deploying the recurrence relation of the binomial coefficients:
    \begin{equation}
		\sum _{k=0}^{B-1} (-1)^k \binom{B+1}{k+1} + 1 = \sum _{k=0}^{B-1} (-1)^k \binom{B}{k} + \sum _{k=0}^{B-1} (-1)^k \binom{B}{k+1} + 1.
	\end{equation}
    But since
    \begin{equation}1= \binom{B}{B}\end{equation}
    we can merge the first and last terms together to get:
  	\begin{equation}
	\begin{aligned}
		& \sum _{k=0}^{B-1} (-1)^k \binom{B}{k} + \sum _{k=0}^{B-1} (-1)^k \binom{B}{k+1} + \binom{B}{B} = \\
		& = \sum _{k=0}^{B} (-1)^k \binom{B}{k} + \sum _{k=0}^{B-1} (-1)^k \binom{B}{k+1}
	\end{aligned}
	\end{equation}
    and using (\ref{PII})
    	\begin{equation}
		\sum _{k=0}^{B} (-1)^k \binom{B}{k} + \sum _{k=0}^{B-1} (-1)^k \binom{B}{k+1} = \sum _{k=0}^{B-1} (-1)^k \binom{B}{k+1}.
	\end{equation}
    Now we perform a change of variable: \(k'=k+1\):
    \begin{equation}
		\sum _{k=0}^{B-1} (-1)^k \binom{B}{k+1} = \sum _{k'=1}^B (-1)^{k'-1} \binom{B}{k'} = -\sum _{k'=1}^B (-1)^{k'} \binom{B}{k'}
	\end{equation}
    and implement the following trick (essentially by adding and subtracting \(1\)):
    \begin{equation}
		-\sum _{k'=1}^B (-1)^{k'} \binom{B}{k'} = -\sum _{k'=1}^B (-1)^{k'} \binom{B}{k'} - \binom{B}{0} + \binom{B}{0}
	\end{equation}
    Now the first two terms of this last expression can be merged together, indeed (\ref{PII}) shows that they mutually cancel out, and we are thus left with:
    	\begin{equation}
		\binom{B}{0}=1
	\end{equation}
    and this concludes our proof for the case \(B\) even; a similar proof holds for the case \(B\) odd. 

    Moving on, we also need to prove equation \ref{second_for_app_b}. This is achieved by expanding it as
    \begin{equation}
		\sum_{k=\rho}^{n}(-1)^k\binom{k}{\rho}\binom{n}{k} = \sum_{k=m}^{n}(-1)^k \frac{k!}{(k-\rho)!\rho!}\frac{n!}{k!(n-k)!}=\binom{n}{\rho} \sum_{k=\rho}^{n}(-1)^k \binom{n-\rho}{k-\rho}
	\end{equation}
    and introducing the variable \(\alpha = k-\rho\):
    \begin{equation}
		\binom{n}{\rho} \sum_{\alpha=0}^{n-\rho}(-1)^{\alpha+\rho} \binom{n-\rho}{\alpha}= (-1)^{\rho} \binom{n}{\rho} \sum_{\alpha=0}^{n-\rho}\binom{n-\rho}{\alpha} (-1)^{\alpha}=0
	\end{equation}
    where the last equality holds thanks to (\ref{PII}).

    \section{More details regarding the first experiment}
    \label{info_experiments}    
\begin{figure}[H]
    \centering
    \includegraphics[width=\textwidth]{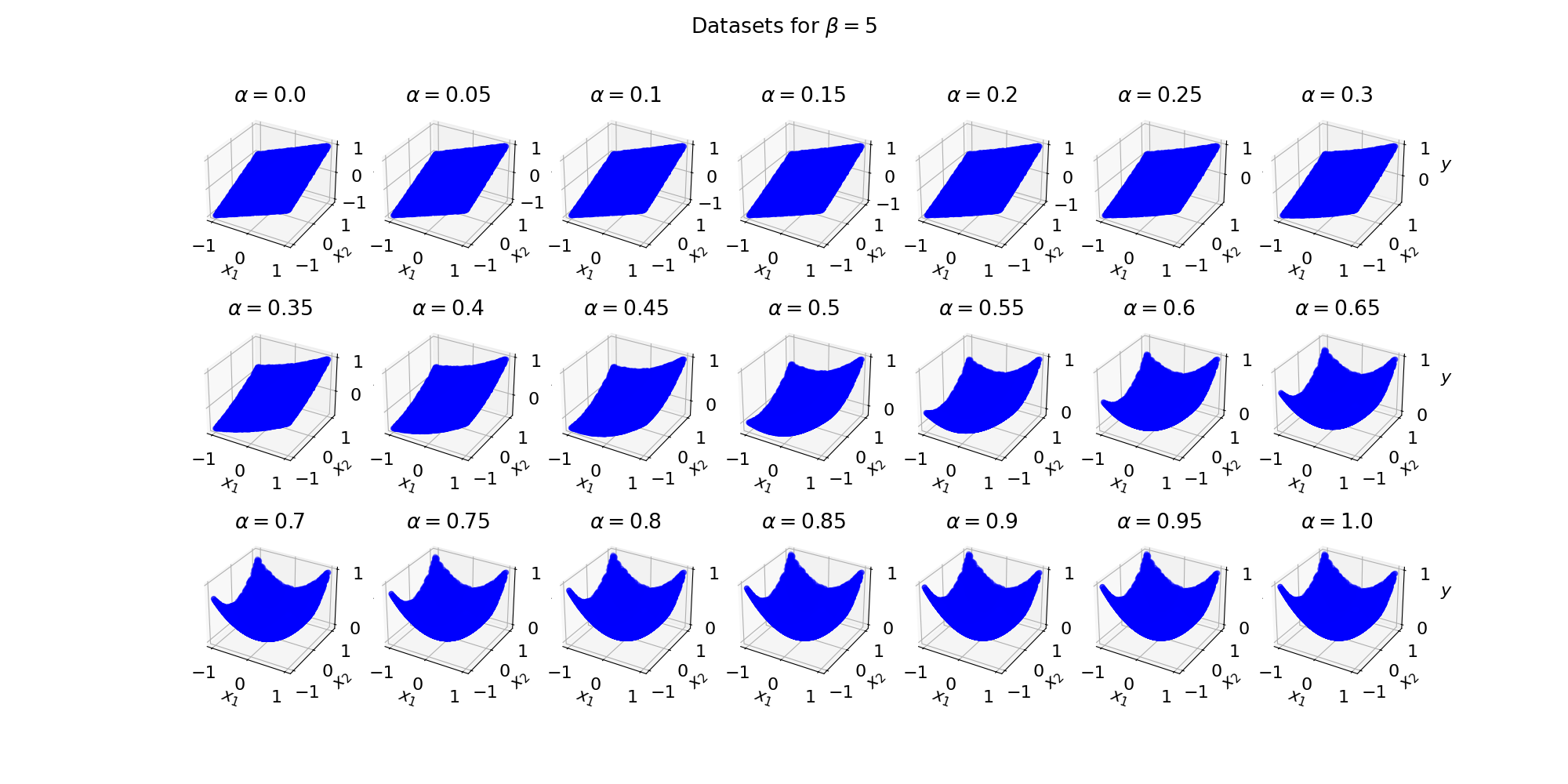}
    \caption{{Dataset for \(\beta = 5\) and \(\alpha\) varying in the range from \(0\) to \(1\).}}
    \label{dataset_figure}
    \end{figure}

    We conducted experiments with a three layer spectral network \(\Phi _2\), with hidden dimension equal to \(300\), and one bias neuron. We trained the network on a multitude of regression problems, with target function \(f(\alpha, \beta; \textbf{x}): \mathbb{R}^2 \to \mathbb{R}\) that becomes progressively non linear with \(\alpha\) (the steepness of the transition is codified by \(\beta\)). In Figure \ref{dataset_figure} the datasets associated with \(\beta = 5\) are shown.

    Different instances of the \(\Phi _2\) model, initialized with only the skip layer connections active, where trained on these datasets; specifically the \(\phi _i\) submatrices were initialized with \textit{Xavier Uniform}\cite{glorot}, and the diagonal elements of \(\mathcal{L}_1,\mathcal{L}_2\) were set to zero, while the elements of the diagonal of \(\mathcal{L}_3\) were set to \(1\).

    We trained the models for \(300\) epochs with a weak \(L_2\) regularization term, acting only on the \(\mathcal{L}_1\) and \(\mathcal{L}_2\) parameters (optimizer being \textit{Adam}, and loss chosen to be the standard \textit{MSE loss}). As expected the resulting models succeed in adequately fitting the datasets.

\begin{figure}[H]
        \centering
        \includegraphics[width=\textwidth]{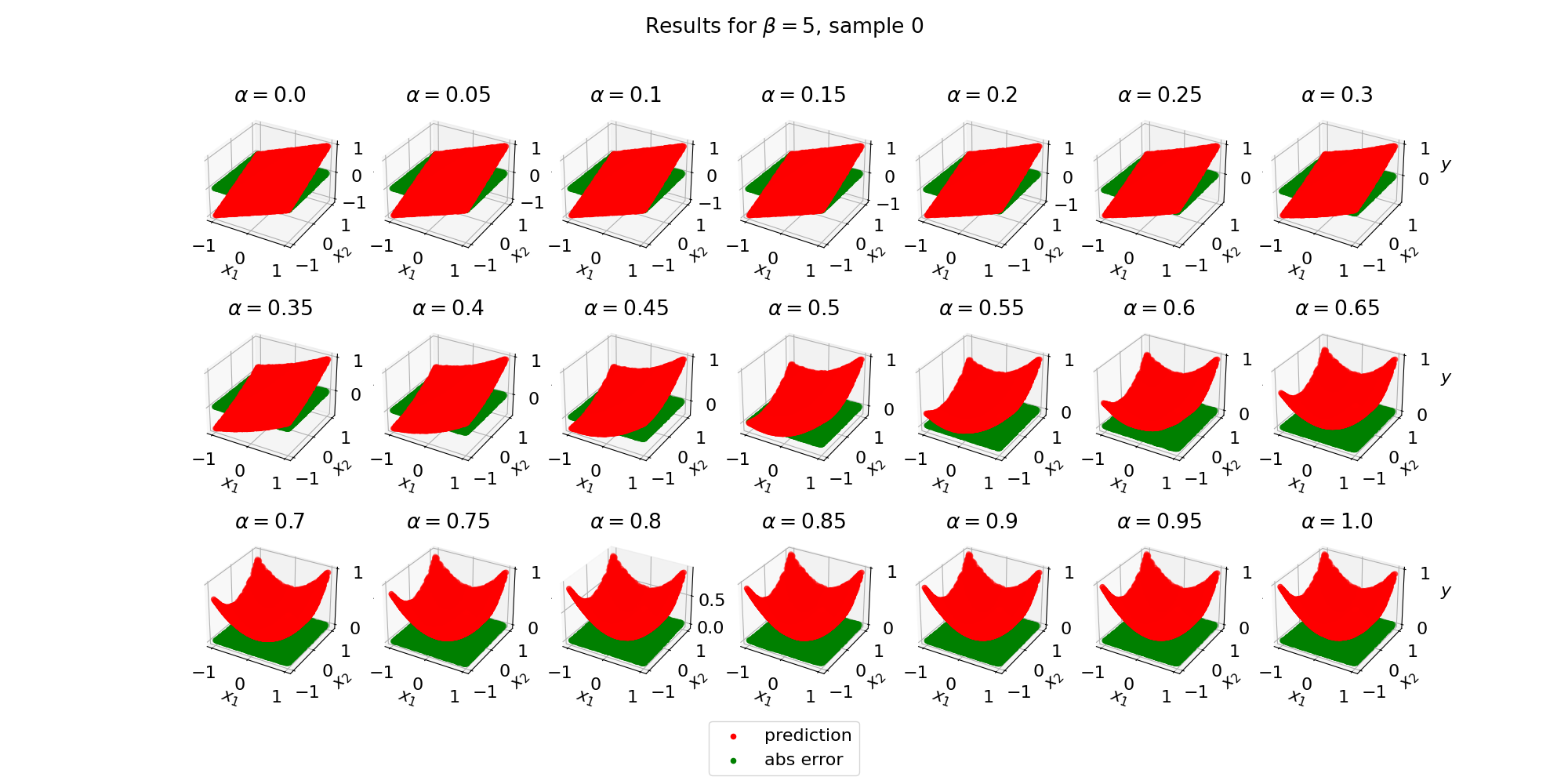}
        \caption{{Fit data for model \(0\) on the datasets with \(\beta = 5\). In red we have shown the model's outputs for different inputs, and in green we report the error associated to the model with respect to the true target function \(f\).}}
        \label{0_goodness_fit}
    \end{figure}

    We trained on two sets of datasets, one with \(\beta = 5\) and one with \(\beta = 10^3\), for a total of \(42\) datasets. For each dataset we trained \(100\) model samples. An example of the goodness of fit for \(\beta = 5\) and model sample number \(0\) is shown in figure \ref{0_goodness_fit}.

    Since we are using the spectral parametrization, the norm of the diagonals of \(\mathcal{L}_1\) and \(\mathcal{L}_2\) provides an indirect measure of the degree of activation of the hidden layer,  the only layer in the network associated with a non-linear step (the \textit{ReLU} activation function).

\begin{figure}[H]
        \centering
        \includegraphics[width=.6\textwidth]{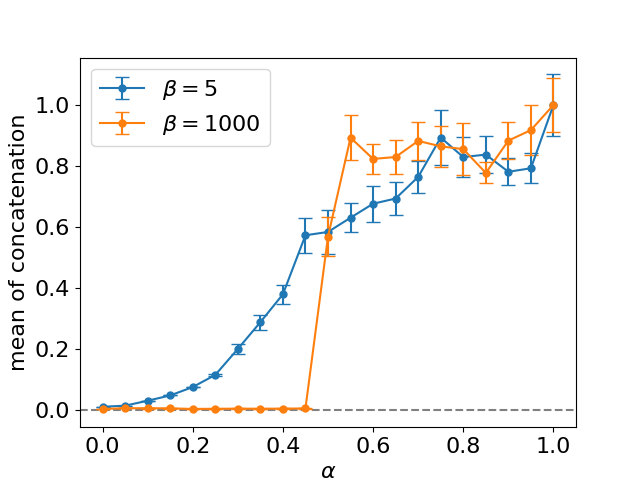}
        \caption{{The mean (on the \(100\) samples) of the concatenation of the \(\mathcal{L}_1,\mathcal{L}_2\) diagonals is plotted against $\alpha$. The data has been normalized before plotting.}}
        \label{L_1L_2}
    \end{figure}

    In Figure \ref{L_1L_2} we show the mean of the norm of the concatenation of \(\mathcal{L}_1,\mathcal{L}_2\) on the model samples for all the different values of \(\alpha, \beta\).

    From Figure \ref{L_1L_2} we can appreciate that the network activates its hidden layer only when needed, namely when non linear terms come into play.

    A clearer picture can be drawn by resorting to a \textit{direct} measure of the network hidden layer contribution to the model's output. To this end we define a tridimensional tensor, called \(\Gamma\), as follows:

    \begin{equation}
        \Gamma_{ijk} = [\mathcal{W}_{32}]_{ij} \cdot [\mathcal{W}_{21}]_{jk}.
    \end{equation}

    Note that there is no summation on the repeated index. This definition ensures that the module of the element \(\Gamma_{ijk}\) measures the strength of the channel that goes from the \(k\)-ith input neuron, pass via the \(j\)-ith hidden neuron, and lands on the \(i\)-ith output neuron. It should be clear that the tensor \(\Gamma\) will associate just one element to each path from the input to the output,  through the hidden (non-linear) layer.

    Given this definition if the norm of \(\Gamma\) is equal to \(0\), no signal is being processed by the hidden layer, and all the information is handled by the linear skip connection layer.

\begin{figure}[ht]
        \centering
        \includegraphics[width=.6\textwidth]{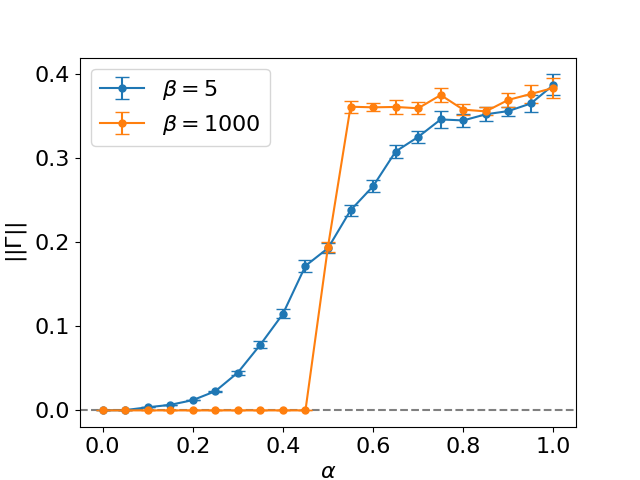}
        \caption{{The normalized square norm of the mean of \(\Gamma\) is plotted versus $\alpha$. Here the mean acts on the model samples.}}
        \label{N_norm}
    \end{figure}

    Figure \ref{N_norm} reports the square norm of the tensor \(\Gamma\) for all the considered datasets.

\begin{figure}[ht]
        \centering
        \includegraphics[width=\textwidth]{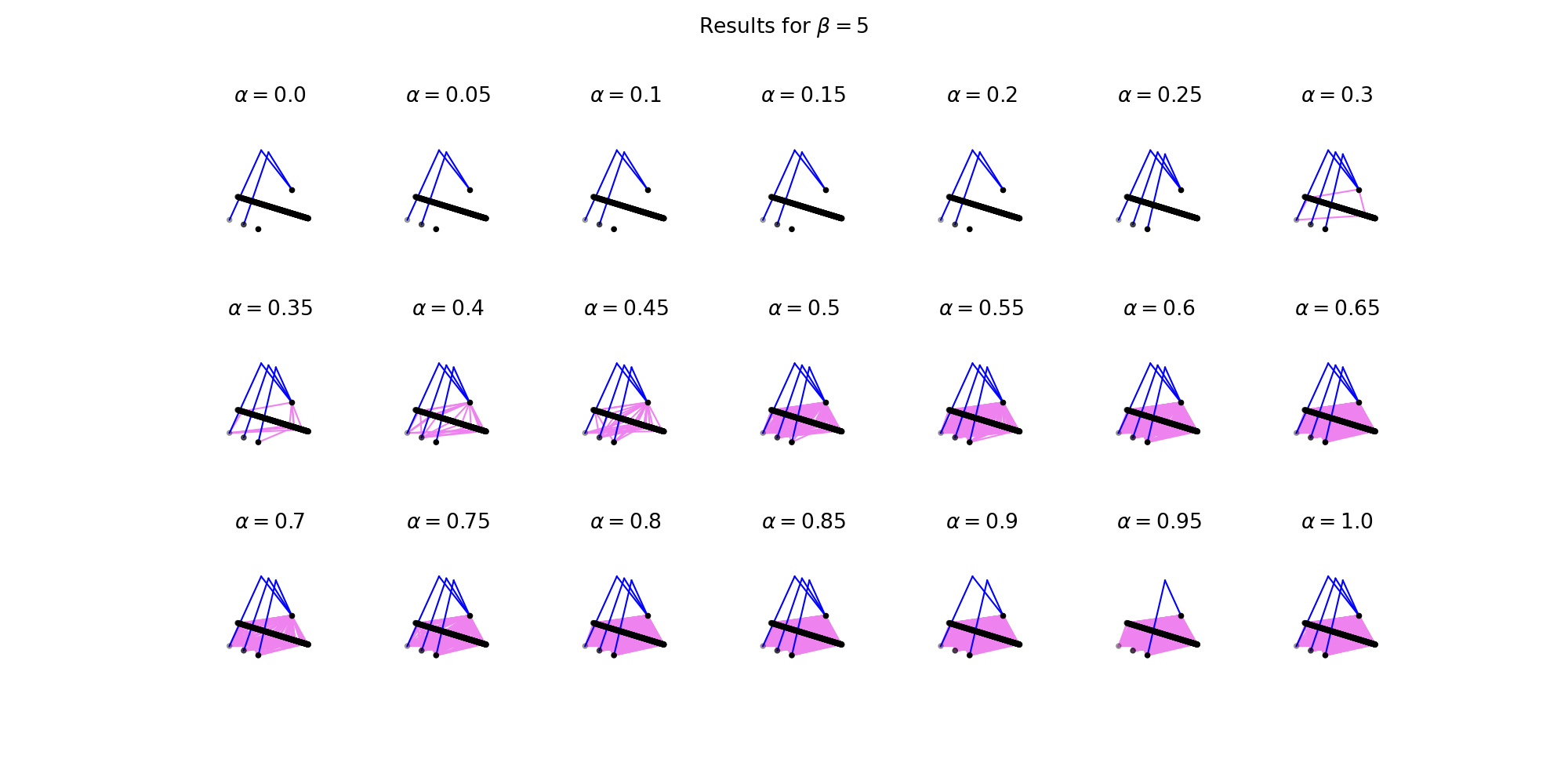}
        \caption{{Draws of the networks' architecture after training via SPARCS. Remember that the training can influence the effective enuing architecture. The networks associated with low $\alpha$ values are effectively two layer networks: the hidden layer is not being used, and can be de facto removed with no impact on the recorded performance.}}
        \label{architectures_sample_0}
    \end{figure}

    The results figure reported in \ref{N_norm} suggest, on the one side, that the spectral parametrization recruits the hidden structures only when needed. Moreover, the activation of the intermediate structures, under weak regularization, is fine tuned to respond with the needed intensity. To better elaborate on this latter point we draw in Figure \ref{architectures_sample_0} the architectures resulting from the training, and depict only the connections with weight larger that an arbitrary cut off set at \(0.01\) for ease of picture interpretability.

    \FloatBarrier

    \section{Additional tests: operating SPARCS in conjunction with Convolutional Neural Networks for image classification tasks}
    \label{AppCIFAR}

To further testing SPARCS, we conducted a series of experiments on benchmark image classification tasks, specifically the CIFAR-10 and CIFAR-100 datasets. The results are reported below. To anticipate the main conclusion, we will prove that SPARCS can effectively operate in conjunction with various complex architectures, trained from scratch or with a fine-tuning of pre-trained models. We would like to point out that all the measurements of the allocated VRAM are reported considering the training phase. During inference the model can be remapped in the direct space, using \eqref{preRelPhiInv} and the amount of memory needed is substantially reduced.
    
    \subsection{Testing on CIFAR-10}

\begin{figure}[ht]
        \centering
        \includegraphics[width=0.8\textwidth]{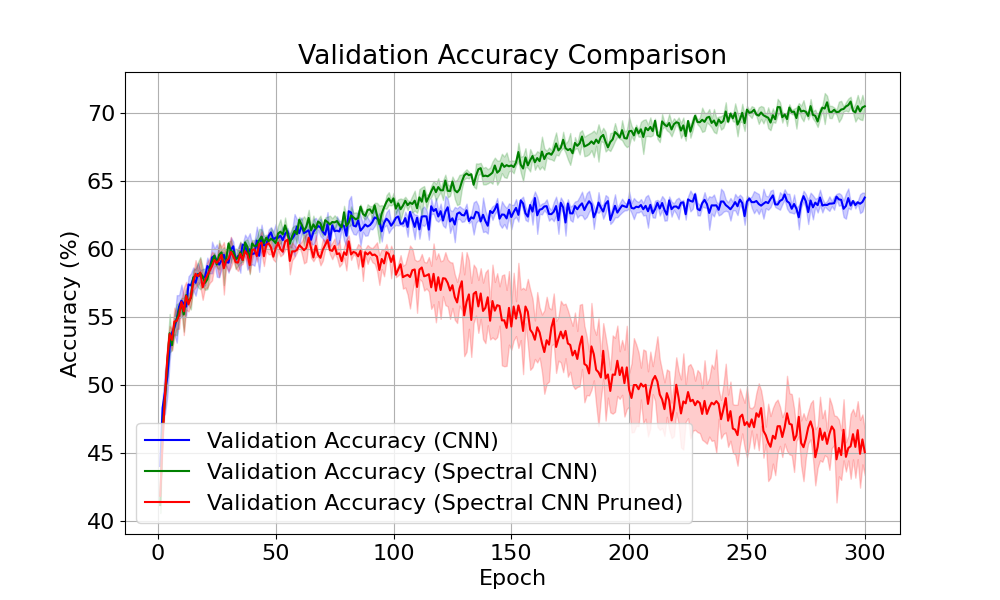}
        \caption{Performance comparison of SPARCS and linear model on CIFAR-10 dataset. The first curve (blue) represents the accuracy of the standard CNN with a linear classifier, while the second curve (green) represents the accuracy of the SPARCS+CNN. The third curve (red) shows the performances of the pruned SPARCS+CNN: this is  obtained by removing all the neurons in the hidden layer. Notice that the three above curves stay close in the initial phases of the training. This seems to imply that the training is initially focused on optimizing the weights in the linear classifier space. However as the training progresses, the SPARCS model begins to potentiate the topology of the network, leading to a significant increase in the recorded performance.}
        \label{fig:cifar10_results}
    \end{figure}

\begin{figure}[ht]
        \centering
        \begin{subfigure}{0.45\textwidth}
            \centering
            \includegraphics[width=\textwidth]{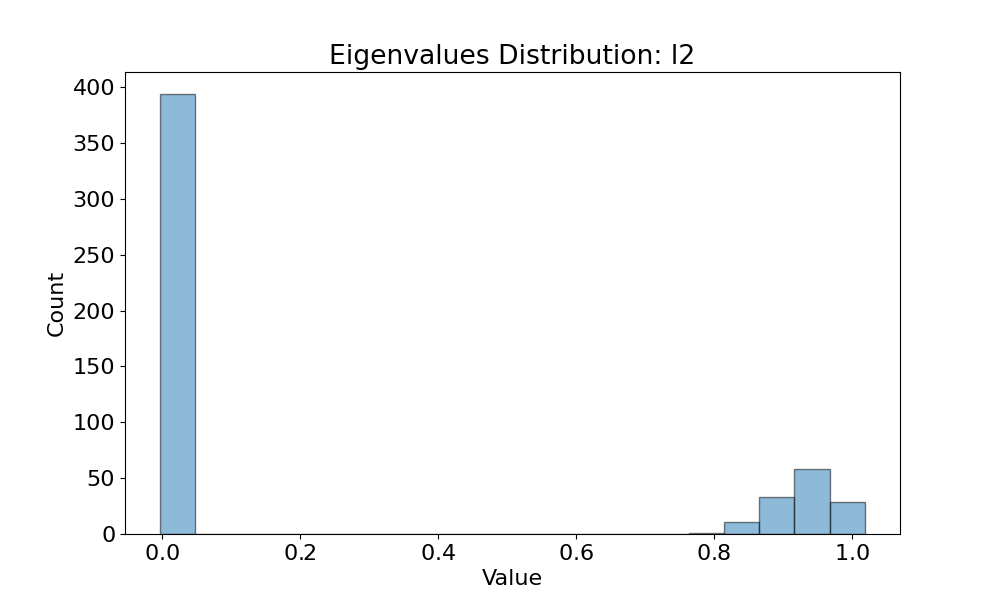}
            \caption{Eigenvalue distribution of the hidden layer in the SPARCS+CNN.}
            \label{fig:eigenvalues_cifar10_hidden}
        \end{subfigure}
        \hfill
        \begin{subfigure}{0.45\textwidth}
            \centering
            \includegraphics[width=\textwidth]{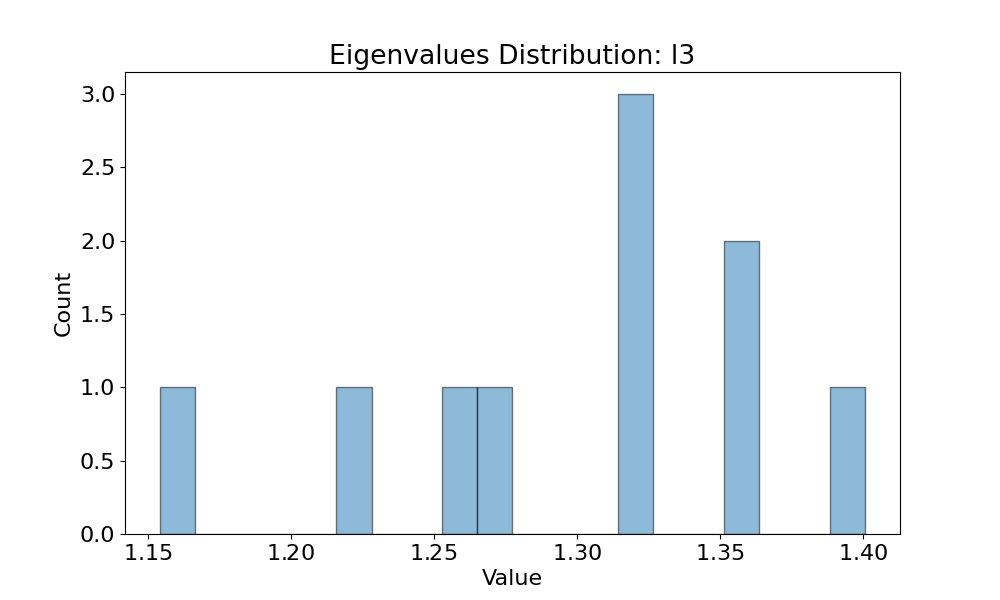}
            \caption{Eigenvalue distribution of the last layer in the SPARCS+CNN.}
            \label{fig:eigenvalues_cifar10_last}
        \end{subfigure}
        \caption{Eigenvalue distributions of the SPARCS+CNN on CIFAR-10 dataset. The left figure shows the eigenvalue distribution of the hidden layer, while the right figure shows the eigenvalue distribution of the last layer. Notice that just a limited fraction of the eigenvalues associated to the hidden layer differ significantly from zero. This amounts to say that the algorithm has spontaneously selected a somehow minimal (or compact) architecture, by recruiting just a finite subset of  active neurons in the hidden layer.}
        \label{fig:eigenvalues_cifar10}
    \end{figure}

The CIFAR-10 dataset consists of 60,000 32x32 color images organized in 10 distinct classes, with 6,000 images per class. The dataset is divided into 50,000 training images and 10,000 test images. To assess the performance of SPARCS in this context we implemented two different models, both equipped with the same minimal convolutional feature extractor block, but with different classification heads. Specifically the feature extractor for both models is composed of convolutional layers, batch normalization, ReLU activation, and max pooling. The first model uses a linear classifier, and is trained in direct space (CNN), while the second model employs a classifier that is initiated as linear, but can grow along training as modulated by SPARCS to become a full nonlinear multi-layer perceptron (SPARCS+CNN). The training is performed from scratch using the Adam optimizer with a learning rate of 0.001, and a batch size of 128. The training is conducted for 300 epochs. The model is tested on the validation set after each epoch. For SPARCS we deploy an eigenvalue regularization constant of \(10^{-4}\), with exclusion of (i) the eigenvalues corresponding to the last (output) layer, which bear no influence on the topology of the network; (ii)  the eigenvalues associated  to the first layer, which are fixed to zero to ensure a one-to-one correspondence between neurons and eigenvalues. The training is performed on a single NVIDIA RTX A5500 GPU, and the memory allocated during training, as well as the time elapsed, are kept track of. The results, shown in Figure \ref{fig:cifar10_results}, indicate that the SPARCS model achieves a higher accuracy as compared to the linear counterpart. This proves again the effectiveness of SPARCS in dynamically evolving the network topology during training so as to improve the recorded performance. This augmented performance is achieved with a modest increase in resource consumption, as shown in table \ref{tab:cifar10_resources}. The eigenvalues corresponding to the neurons in the hidden layer, initially set to zero, grow to progressively assume values of the same order of magnitude as that displayed by the eigenvalues associated to the last layer of the collection. In words, SPARCS is capable of  adjusting its topology to better tackle the supplied classification task. Interestingly, we observe the formation of two clusters in the eigenvalues distribution across the hidden layer: a limited fraction of the neurons are thus selected for usage, so yielding a compact architecture that will have reduced memory footprint and low computational cost in deployment and exploitation. The eigenvalue distribution is shown in Figure \ref{fig:eigenvalues_cifar10}.
    
    \begin{table}[ht]
        \centering
        \begin{tabular}{|c|c|c|c|}
            \hline
            Model & GPU\tablefootnote{Memory allocated during training.} (MB) & Training Time (s) & Test Accuracy (\%) \\
            \hline
            CNN & 17.9 & 1357.0 & 68.2 \\
            SPARCS+CNN & 21.2 & 1361.98  & 73.9 \\
            \hline
        \end{tabular}
        \caption{Comparison of resource consumption and performance resulting from the training of the SPARCS+CNN and the standard CNN on CIFAR-10 dataset. Notice how the SPARCS model achieves a significantly higher accuracy with only a modest increase in GPU memory consumption and training time.}
        \label{tab:cifar10_resources}
    \end{table}
    
    \FloatBarrier
    
    \subsection{Testing on CIFAR-100}
    
    The CIFAR-100 has a similar structure to CIFAR-10, but it contains 100 classes with 600 images per class. In what follows, the same experimental setup as for CIFAR-10 is used, except for the adoption of a powerful pre-trained CNN architecture. Although the architecture used after the backbone is comparable to the previous one, this test shows the usability of SPARCS in realistic contexts and tasks, and its ability to operate with the data distribution downstream of a feature extraction. Tests integrating the method inside the full-architecture, with consequent `from scratch' training, are left or upcoming work.
    The employed backbone is  EfficientNetV2-S, a competitive convolutional neural network architecture. For both examined models, we fine-tune the pre-trained EfficientNetV2-S model, upon replacing the embedded output classifier with (i) a new linear module for the reference benchmark model and (ii) a SPARCS classifier for the competitor scheme. The fine tuning (one pass through the CIFAR-100 training set) is performed with the same hyperparameters employed for CIFAR-10, and in particular with the same regularization constant set to \(10^{-4}\). Given the heavy amount of preprocessing going on in EfficientNetV2-S (and also since the original state-of-the-art like EfficientNetV2-S model is already equipped with a linear classifier), one can guess that the potential benefits as stemming by the activation of just one additional non linear layer in the classifier SPARCS section, would be negligible. 
    Remarkably SPARCS, operated only on the train set, automatically finds that there is no need to augment the topology of the network to cope with the assigned task, given the massive pre-processing of the data. Hence, the linear initialization is kept unchanged through SPARCS. In table \ref{tab:cifar100_resources} we show the resource consumption and performance of the SPARCS EfficientNetV2-S model as compared to the standard EfficientNetV2-S model.

    \begin{table}[ht]
        \centering
        \begin{tabular}{|c|c|c|c|}
            \hline
            Model & GPU\tablefootnote{Memory allocated during training.} (MB) & Fine Tuning Time (s) & Test Accuracy (\%) \\
            \hline
            EffNet & 408.4 & 138.8 & 73.2 \\
            SPARCS EffNet & 518.5 & 139.5 & 72.7 \\
            \hline
        \end{tabular}
        \caption{Comparison of resource consumption  and performance resulting from the fine-tuning of EfficientNetV2-S (EffNet) to CIFAR-100. Notice how both fine tuning paradigms achieve a similar performance: no advantage is found in recruiting more neurons from the hidden layer in the SPARCS classifier module.}
        \label{tab:cifar100_resources}
    \end{table}
    
    \begin{figure}[ht]
        \centering
        \begin{subfigure}{0.45\textwidth}
            \centering
            \includegraphics[width=\textwidth]{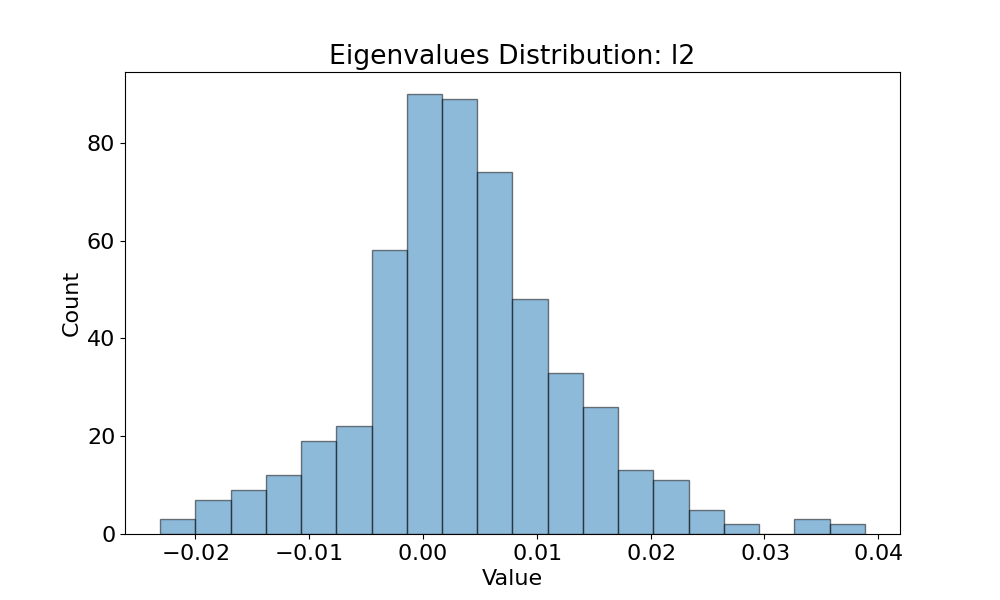}
            \caption{Eigenvalues of the hidden layer in the SPARCS EfficientNetV2-S.}
            \label{fig:eigenvalues_cifar100_hidden}
        \end{subfigure}
        \hfill
        \begin{subfigure}{0.45\textwidth}
            \centering
            \includegraphics[width=\textwidth]{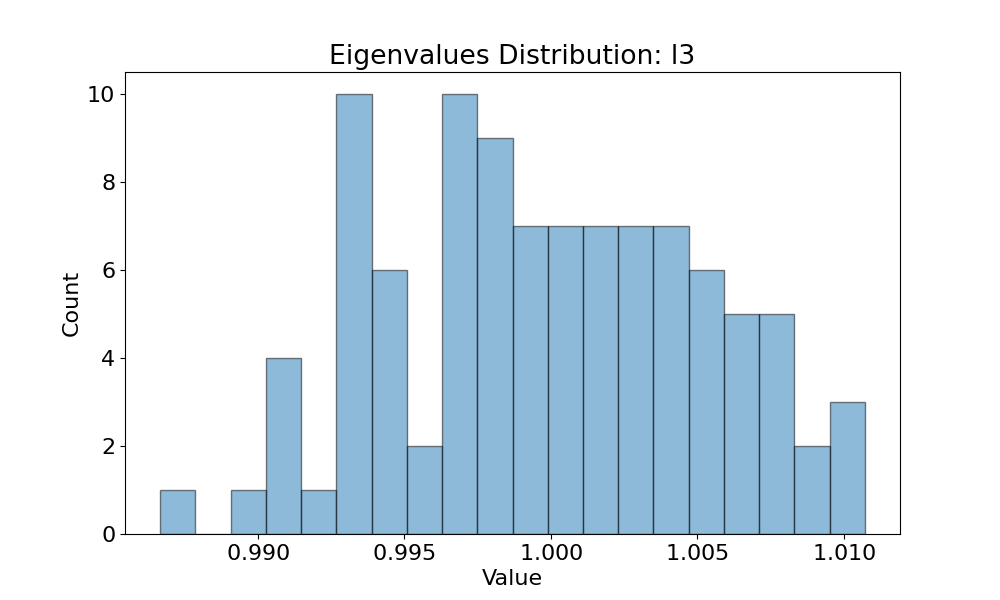}
            \caption{Eigenvalue distribution of the last layer in the SPARCS EfficientNetV2-S.}
            \label{fig:eigenvalues_cifar100_last}
        \end{subfigure}
        \caption{Eigenvalue distributions of the SPARCS EfficientNetV2-S on CIFAR-100 dataset. The left figure shows the eigenvalue distribution of the hidden layer, while the right figure shows the eigenvalue distribution of the last layer. Notice how the eigenvalues of the hidden layer are all orders of magnitude smaller than the eigenvalues of the last layer. This implies that SPARCS model did not call for an augmented topology, as it could have been anticipated given that the EfficientNetV2-S is already complex enough for handling the CIFAR-100 dataset.}
        \label{fig:eigenvalues_cifar100}
    \end{figure}

    \FloatBarrier

    \section{Hyperparameters used}
    \label{hyperparameters}

For what concerns the regularization, in the experiments carried out, we have used both L1 and L2 norms as weight penalties. In order to further asses the flexibility of the method, we also varied the regularization strength parameter, and SPARCS appears to work well for a reasonable wide range of values (ranging approximately from \(10^{-3}\) to \(10^{-5}\)). For many tasks however the sweet spot seem to be located around \(10^{-4}\), following heuristics that its value should be around \(10^{-3}\) of the Data Loss. In the following we present the parameters used for each task in the associated table.

\newpage

    \begin{center}
        \textbf{Hyperparameters for the Simple Regression Experiment}
    \end{center}

    \begin{table}[H]
        \centering
        \begin{tabular}{@{}l c@{}}
            \hline
            \textbf{Hyperparameter} & \textbf{Value} \\
            \hline
            \rule{0pt}{2.5ex}Learning Rate & \(10^{-3}\) \\
            Batch Size & 100 \\
            Number of Epochs & 300 \\
            Max Number of Hidden Neurons & 300 \\
            Optimizer & Adam \\
            Regularization Type & L2 \\
            Regularization Strength & \(10^{-4}\) \\
            Activation Function & ReLU \\
            Loss Function & MSE \\
            NumPy Seed & 42 \\
            Number of Trials & 100 \\
            \hline
        \end{tabular}
        \caption{Hyperparameters used in the first experiment. The numpy seed is used in the data generation phase, while \textit{number of trials} refers to the number of repetition needed to give an estimate of the error bars in the results (standard deviations).}
    \end{table}

    \begin{center}
        \textbf{Hyperparameters for the Teacher-Student Experiment.}
    \end{center}

    \begin{table}[H]
        \centering
        \begin{tabular}{@{}l c@{}}
            \hline
            \textbf{Hyperparameter} & \textbf{Value} \\
            \hline
            \rule{0pt}{2.5ex}Learning Rate & \(10^{-3}\) \\
            Batch Size & 1024 \\
            Number of Epochs & 180 \\
            Number of Neurons in Teacher Hidden Layer & 20 \\
            Max Number of Neurons in Student Hidden Layers & 200 \\
            Optimizer & Adam \\
            Regularization Type & L2 \\
            Regularization Strength & \(3 \cdot 10^{-3}\) \\
            Activation Function & ReLU \\
            Loss Function & MSE \\
            NumPy Seed & 42 \\
            \hline
        \end{tabular}
        \caption{Hyperparameters used in the second experiment. The numpy seed is used in the data generation phase.}
    \end{table}

    \newpage
    
    \begin{center}
        \textbf{Hyperparameters for the CIFAR 10 and CIFAR 100 Experiments}
    \end{center}

    \begin{table}[H]
        \centering
        \begin{tabular}{@{}l c@{}}
            \hline
            \textbf{Hyperparameter} & \textbf{Value} \\
            \hline
            \rule{0pt}{2.5ex}Learning Rate & \(10^{-3}\) \\
            Batch Size & 128 \\
            Number of Epochs & 300 \\
            Max Number of Hidden Neurons & 526 \\
            Optimizer & Adam \\
            Regularization Type & L1 \\
            Regularization Strength & \(10^{-4}\) \\
            Activation Function & ReLU \\
            Loss Function & Log Loss \\
            Number of Trials & 100 \\
            \hline
        \end{tabular}
        \caption{Hyperparameters used in the third and fourth experiments (in appendix). Notice that all the parameters have been kept equal between the two experiments, except for the number of epochs: since on CIFAR 100 we only fine tuned, instead of training from scratch, we performed a single pass trough the training data. Also notice that with the term \textit{log loss} we of course refer to the standard Categorical Cross-Entropy. Once again the \textit{number of trials} line refers to the error estimation via empirical standard deviation measurements.}
    \end{table}

\FloatBarrier
	
\bibliographystyle{unsrt}  
\bibliography{bibliography.bib}  
	
\end{document}

%% file: Phi_1.tex
\begin{tikzpicture}[scale=0.8]


\draw[thick] (0,0) rectangle (6,6);
\node at (-1,3) {\Large $\Phi=$};


\draw (0,6) -- (6,0);


\draw (0,0) rectangle (4,2) node [pos=.5] {\large $\phi$};





\draw [decorate,decoration={brace,amplitude=10pt},xshift=0pt,yshift=4pt]
(0,6) -- (4,6) node [black,midway,yshift=0.6cm] {$N_1$};

\draw [decorate,decoration={brace,amplitude=10pt},xshift=0pt,yshift=4pt]
(4,6) -- (6,6) node [black,midway,yshift=0.6cm] {$N_2$};


\draw [decorate,decoration={brace,amplitude=10pt},xshift=4pt,yshift=0pt]
(6,6) -- (6,2) node [black,midway,xshift=0.6cm] {$N_1$};

\draw [decorate,decoration={brace,amplitude=10pt},xshift=4pt,yshift=0pt]
(6,2) -- (6,0) node [black,midway,xshift=0.6cm] {$N_2$};

\end{tikzpicture}

%% file: Lambda_1.tex
\begin{tikzpicture}[scale=0.8]

\tikzset{decoration={snake,amplitude=.4mm,segment length=2mm, post length=0mm,pre length=0mm}}


\draw[thick] (0,0) -- (0,6) -- (6,6) -- (6,0) -- (0,0);


\draw[decorate] (0,6) -- (6,0);

\draw[dashed] (0,2) -- (6,2);
\draw[dashed] (4,0) -- (4,6);

\node at (2.5,4.25) {\large$\mathcal{L}_1$}; 
\node at (5.5,1.25) {\large$\mathcal{L}_2$}; 

\node at (-1,3) {\large$\Lambda =$};





\draw [decorate,decoration={brace,amplitude=10pt},xshift=0pt,yshift=4pt]
(0,6) -- (4,6) node [black,midway,yshift=0.6cm] {$N_1$};

\draw [decorate,decoration={brace,amplitude=10pt},xshift=0pt,yshift=4pt]
(4,6) -- (6,6) node [black,midway,yshift=0.6cm] {$N_2$};


\draw [decorate,decoration={brace,amplitude=10pt},xshift=4pt,yshift=0pt]
(6,6) -- (6,2) node [black,midway,xshift=0.6cm] {$N_1$};

\draw [decorate,decoration={brace,amplitude=10pt},xshift=4pt,yshift=0pt]
(6,2) -- (6,0) node [black,midway,xshift=0.6cm] {$N_2$};

\end{tikzpicture}

%% file: A_1.tex
\begin{tikzpicture}[scale=0.8]

\tikzset{decoration={snake,amplitude=.4mm,segment length=2mm, post length=0mm,pre length=0mm}}

\draw[thick] (0,0) -- (0,6) -- (6,6) -- (6,0) -- (0,0);

\draw[decorate] (0,6) -- (6,0);

\draw (0,0) rectangle (4,2) node [pos=.5] {\large $\phi \mathcal{L}_1-\mathcal{L}_2\phi $};

\draw[dashed] (0,2) -- (6,2);
\draw[dashed] (4,0) -- (4,6);

\node at (2.5,4.25) {\large$\mathcal{L}_1$}; 
\node at (5.5,1.25) {\large$\mathcal{L}_2$}; 

\node at (-1.2,3) {\large$A =$};





\draw [decorate,decoration={brace,amplitude=10pt},xshift=0pt,yshift=4pt]
(0,6) -- (4,6) node [black,midway,yshift=0.6cm] {$N_1$};

\draw [decorate,decoration={brace,amplitude=10pt},xshift=0pt,yshift=4pt]
(4,6) -- (6,6) node [black,midway,yshift=0.6cm] {$N_2$};


\draw [decorate,decoration={brace,amplitude=10pt},xshift=4pt,yshift=0pt]
(6,6) -- (6,2) node [black,midway,xshift=0.6cm] {$N_1$};

\draw [decorate,decoration={brace,amplitude=10pt},xshift=4pt,yshift=0pt]
(6,2) -- (6,0) node [black,midway,xshift=0.6cm] {$N_2$};

\end{tikzpicture}

%% file: Phi_2.tex
\begin{tikzpicture}[scale=1.2]


	\draw[thick] (0,0) rectangle (6,6);
	\node at (-1,3) {\Large $\Phi_2=$};


	\draw (0,6) -- (6,0);


	\draw (0,4) rectangle (2,1) node [pos=.5] {\large $\phi_1^{(2)}$};
	\draw (2,0) rectangle (5,1) node [pos=.5] {\large $\phi_2^{(2)}$};



	\draw [decorate,decoration={brace,amplitude=10pt},xshift=0pt,yshift=4pt]
	(0,6) -- (2,6) node [black,midway,yshift=0.6cm] {$N_1$};

	\draw [decorate,decoration={brace,amplitude=10pt},xshift=0pt,yshift=4pt]
	(2,6) -- (5,6) node [black,midway,yshift=0.6cm] {$N_2$};

	\draw [decorate,decoration={brace,amplitude=10pt},xshift=0pt,yshift=4pt]
	(5,6) -- (6,6) node [black,midway,yshift=0.6cm] {$N_3$};


	\draw [decorate,decoration={brace,amplitude=10pt},xshift=4pt,yshift=0pt]
	(6,6) -- (6,4) node [black,midway,xshift=0.6cm] {$N_1$};

	\draw [decorate,decoration={brace,amplitude=10pt},xshift=4pt,yshift=0pt]
	(6,4) -- (6,1) node [black,midway,xshift=0.6cm] {$N_2$};

	\draw [decorate,decoration={brace,amplitude=10pt},xshift=4pt,yshift=0pt]
	(6,1) -- (6,0) node [black,midway,xshift=0.6cm] {$N_3$};

\end{tikzpicture}

%% file: Lambda_2.tex
\begin{tikzpicture}[scale=1.2]

\tikzset{decoration={snake,amplitude=.4mm,segment length=2mm, post length=0mm,pre length=0mm}}


\draw[thick] (0,0) -- (0,6) -- (6,6) -- (6,0) -- (0,0);


\draw[decorate] (0,6) -- (6,0);

\draw[dashed] (0,6) rectangle (2,4);
\draw[dashed] (2,4) rectangle (5,1);
\draw[dashed] (5,1) rectangle (6,0);

\node at (1.5,5.25) {\large$\mathcal{L}_1^{(2)}$};
\node at (4,3) {\large$\mathcal{L}_2^{(2)}$};
\node at (5.7,0.7) {\large$\mathcal{L}_3^{(2)}$}; 
\draw [decorate,decoration={brace,amplitude=10pt},xshift=0pt,yshift=4pt]
		(0,6) -- (2,6) node [black,midway,yshift=0.6cm] {$N_1$};

\draw [decorate,decoration={brace,amplitude=10pt},xshift=0pt,yshift=4pt]
		(2,6) -- (5,6) node [black,midway,yshift=0.6cm] {$N_2$};

\draw [decorate,decoration={brace,amplitude=10pt},xshift=0pt,yshift=4pt]
		(5,6) -- (6,6) node [black,midway,yshift=0.6cm] {$N_3$};

\draw [decorate,decoration={brace,amplitude=10pt},xshift=4pt,yshift=0pt]
		(6,6) -- (6,4) node [black,midway,xshift=0.6cm] {$N_1$};

		\draw [decorate,decoration={brace,amplitude=10pt},xshift=4pt,yshift=0pt]
		(6,4) -- (6,1) node [black,midway,xshift=0.6cm] {$N_2$};

		\draw [decorate,decoration={brace,amplitude=10pt},xshift=4pt,yshift=0pt]
		(6,1) -- (6,0) node [black,midway,xshift=0.6cm] {$N_3$};

\node at (-1,3) {\Large $\Lambda_2 =$};

\end{tikzpicture}

%% file: A_2.tex
\begin{tikzpicture}[scale=1.6]

\tikzset{decoration={snake,amplitude=.4mm,segment length=2mm, post length=0mm,pre length=0mm}}


\draw[thick] (0,0) -- (0,6) -- (6,6) -- (6,0) -- (0,0);


\draw[decorate] (0,6) -- (6,0);

\draw[dashed] (0,6) rectangle (2,4);
\draw[dashed] (2,4) rectangle (5,1);
\draw[dashed] (5,1) rectangle (6,0);

\node at (1.5,5.25) {\large$\mathcal{L}_1^{(2)}$};
\node at (4,3) {\large$\mathcal{L}_2^{(2)}$};
\node at (5.7,0.7) {\large$\mathcal{L}_3^{(2)}$}; 
\draw [decorate,decoration={brace,amplitude=10pt},xshift=0pt,yshift=4pt]
		(0,6) -- (2,6) node [black,midway,yshift=0.6cm] {$N_1$};

\draw [decorate,decoration={brace,amplitude=10pt},xshift=0pt,yshift=4pt]
		(2,6) -- (5,6) node [black,midway,yshift=0.6cm] {$N_2$};

\draw [decorate,decoration={brace,amplitude=10pt},xshift=0pt,yshift=4pt]
		(5,6) -- (6,6) node [black,midway,yshift=0.6cm] {$N_3$};

\draw [decorate,decoration={brace,amplitude=10pt},xshift=4pt,yshift=0pt]
		(6,6) -- (6,4) node [black,midway,xshift=0.6cm] {$N_1$};

		\draw [decorate,decoration={brace,amplitude=10pt},xshift=4pt,yshift=0pt]
		(6,4) -- (6,1) node [black,midway,xshift=0.6cm] {$N_2$};

		\draw [decorate,decoration={brace,amplitude=10pt},xshift=4pt,yshift=0pt]
		(6,1) -- (6,0) node [black,midway,xshift=0.6cm] {$N_3$};

\draw (0,4) rectangle (2,1);
\draw (2,0) rectangle (6,1);

\node at (1,2.5) {\small$\phi ^{(2)}_1\mathcal{L}^{(2)}_1-\mathcal{L}^{(2)}_2\phi _1^{(2)}$};
\node at (3.5,0.5) {\small$\phi ^{(2)}_2\mathcal{L}^{(2)}_2-\mathcal{L}^{(2)}_3\phi _2^{(2)}$};
\node at (1,0.5) {\tiny $\Big(\mathcal{L}^{(2)}_3\phi ^{(2)}_2-\phi ^{(2)}_2 \mathcal{L}^{(2)}_2\Big)\phi ^{(2)}_1$};

\node at (-0.5,3) {\large $A_2 =$};

\end{tikzpicture}